\documentclass[10pt,twocolumn,letterpaper]{article}
\usepackage{cvpr}
\usepackage{times}
\usepackage{epsfig}
\usepackage{graphicx}
\usepackage{amsmath}
\usepackage{amssymb}
\usepackage{float}
\usepackage{amsmath,diagbox,tabu,stackengine}
\usepackage[]{algorithm2e}
\usepackage{url}            
\usepackage{booktabs}       
\usepackage{amsfonts}       

\usepackage{caption}
\usepackage{multirow}
\usepackage{xspace}
\usepackage{mathtools}
\usepackage{wasysym}
\usepackage{marvosym}
\usepackage[table,xcdraw]{xcolor}
\usepackage{pdfcomment}
\usepackage{lipsum}
\usepackage{subcaption}
\usepackage{placeins}
\usepackage{hyperref}
\hypersetup{breaklinks=true,colorlinks}

\cvprfinalcopy 

\def\cvprPaperID{2599} 

\ifcvprfinal\fi

\def\by{\mathbf y}
\def\byhat{{\mathbf{\hat{y}}}}

\makeatother

\title{\vspace{-4mm} Devil is in the Edges: Learning Semantic Boundaries from Noisy Annotations}

\author{
	David Acuna$^{1,2,3}$
	\hspace{2cm}
	Amlan Kar$^{2,3}$  \hspace{2cm}
	Sanja Fidler$^{1,2,3}$\\
$^1$NVIDIA \hspace{2em} $^2$University of Toronto \hspace{2em}   $^3$Vector Institute  \\
{\tt\small \{davidj, amlan\}@cs.toronto.edu}, {\tt\small sfidler@nvidia.com} 
}

\begin{document}

\maketitle

\begin{abstract}
We tackle the problem of semantic boundary prediction, which aims to identify pixels that belong to object(class) boundaries. We notice that relevant datasets consist of a significant level of label noise, reflecting the fact that precise annotations are laborious to get and thus annotators trade-off quality with efficiency. We aim to learn sharp and precise semantic boundaries by explicitly reasoning about annotation noise during training. We propose a simple new layer and loss that can be used with existing learning-based boundary detectors. Our layer/loss enforces the detector to predict a maximum response along the normal direction at an edge, while also regularizing its direction. 
We further reason about true object boundaries during training using a level set formulation, which allows the network to learn from misaligned labels in an end-to-end fashion. Experiments show that we improve over the CASENet~\cite{yu2017casenet} backbone network by more than 4\% in terms of MF(ODS) and 18.61\% in terms of AP, outperforming all current state-of-the-art methods including those that deal with alignment. Furthermore, we show that our learned network can be used to significantly improve coarse segmentation labels, lending itself as an efficient way to label new data. \\\small{Project Page: \href{https://nv-tlabs.github.io/STEAL/}{https://nv-tlabs.github.io/STEAL/}}

\end{abstract}

\vspace{-3mm}
\section{Introduction}
\label{sec:intro}

%

Image boundaries are an important cue for recognition~\cite{Opelt:2006,ShapeHierarchy,amfm_pami2011}. Humans can recognize objects from sketches alone, even in cases where a significant portion of the boundary is missing~\cite{Biederman87,getsalt}. Boundaries have also been shown to be useful for 3D reconstruction~\cite{MalikM89,Lee2009GeometricRF,disp}, localization~\cite{VLASE,WangICCV15}, and image generation~\cite{pix2pix2016,wang2018pix2pixHD}.

In the task of semantic boundary detection, the goal is to move away from low-level image edges to identifying image pixels that belong to object (class) boundaries. 
It can be seen as a dual task to image segmentation which identifies object regions. 
Intuitively, predicting semantic boundaries is an easier learning task since they are mostly rooted in identifiable higher-frequency image locations, while region pixels may often be homogenous in color, leading to ambiguities for recognition. On the other hand, the performance metrics are harder: while getting the coarse regions right %
may lead to artificially high Jaccard index~\cite{Geodesic14}, boundary-related metrics focus their evaluation tightly along the object edges. Getting these correct is very important for tasks such as object instance segmentation, robot manipulation and grasping, or image editing.

\begin{figure}[t!]
\vspace{-2mm}
\includegraphics[width=\linewidth,trim=0 0 0 0,clip]{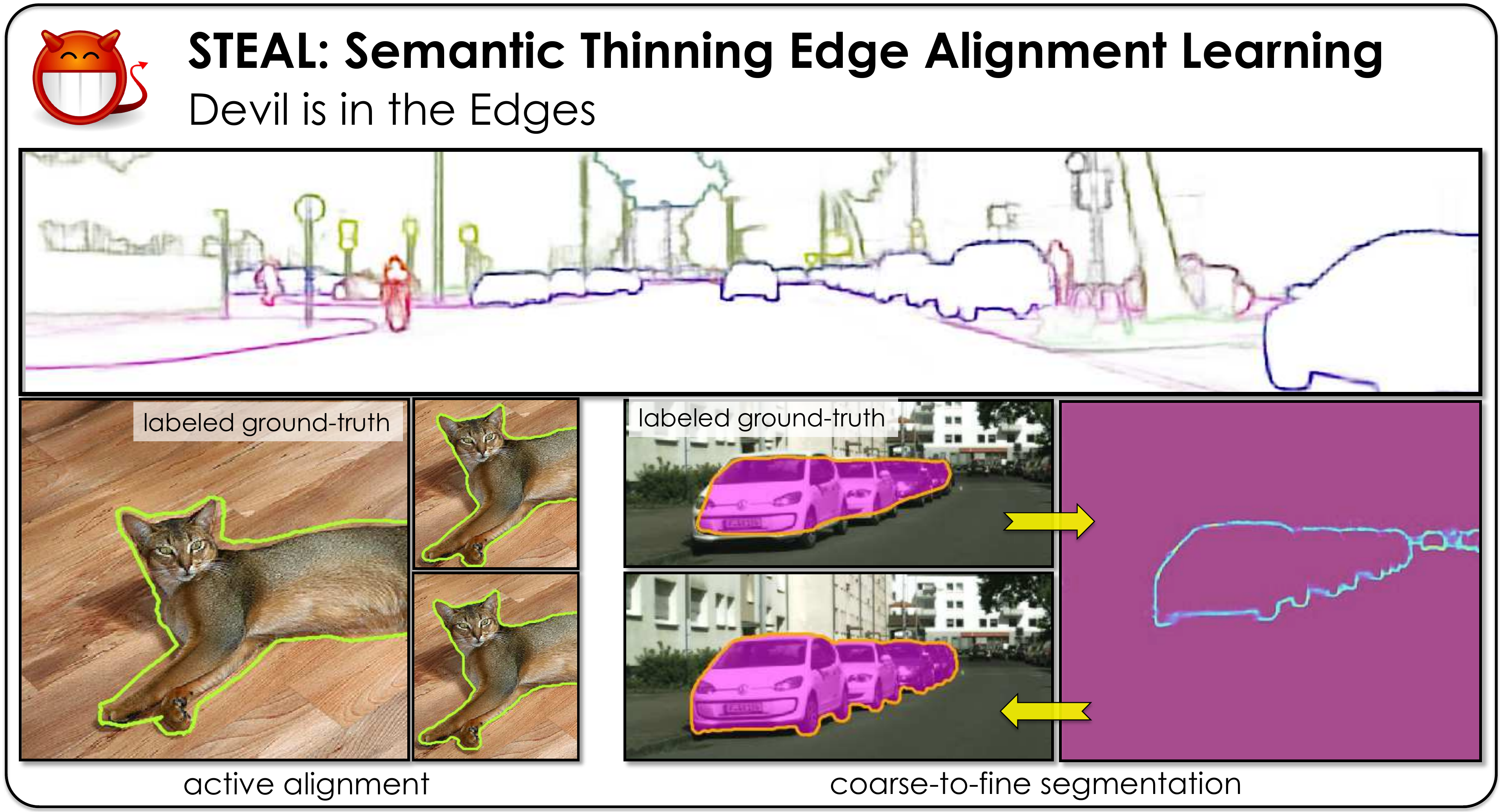} 
\vspace{-7mm}
\caption{We introduce STEAL, an approach to learn sharper and more accurate semantic boundaries. STEAL can be plugged onto any existing semantic boundary network, and is able to significantly refine noisy annotations in current datasets.}
\label{fig:intro}
\vspace{-3mm}
\end{figure}

However, annotating precise object boundaries is extremely slow, taking as much as 30-60s per object~\cite{PolygonPP2018,ChenCVPR14}. Thus most existing datasets consist of significant label noise (Fig.~\ref{fig:intro}, bottom left), trading quality with the labeling efficiency. This may be the root cause why most learned detectors output thick boundary predictions, which are undesirable for downstream tasks.

In this paper, we aim to learn sharp and precise semantic boundaries by explicitly reasoning about annotation noise during training. We propose a new layer and loss that can be added on top of any end-to-end edge detector. It enforces the edge detector to predict a maximum response along the normal direction at an edge, while also regularizing its direction. By doing so, we alleviate the problem of predicting overly thick boundaries and directly optimize for Non-Maximally-Suppressed (NMS) edges. We further reason about true object boundaries using a level-set formulation, which allows the network to learn from misaligned labels in an end-to-end fashion. 

Experiments show that our approach improves the performance of a backbone network, \ie CASENet~\cite{yu2017casenet}, by more than 4\% in terms of MF(ODS) and 18.61\% in terms of AP, outperforming all current state-of-the-art methods. We further show that our predicted boundaries are significantly better than those obtained from the latest DeepLab-v3~\cite{deeplabv3plus2018} segmentation outputs, while using a much more lightweight architecture. Our learned network is also able to improve coarsely annotated segmentation masks with 16px, 32px error improving their accuracy by more than 20\% IoU and 30\% IoU, respectively.  This lends our method as an efficient means to collect new labeled data, allowing annotators to coarsely outline  objects with just a few clicks, and generating finer ground-truth using our approach. We showcase this idea by refining the Cityscapes-coarse labelset, and exploiting these labels to train a state-of-the-art segmentation network~\cite{deeplabv3plus2018}. We observe a significant improvement of more than 1.2\% in some of the refined categories.

\vspace{-1mm}
\section{Related Work}
\label{sec:related}
 \vspace{-1mm}

 {\bf Semantic Boundary Detection.}
Learning-based semantic edge detection dates back to~\cite{prasad2006learning} which learned a classifier that operates on top of a standard edge detector. 
In~\cite{BharathICCV2011}, the authors introduced the Semantic Boundaries Dataset (SBD) and formally studied the problem of semantic contour detection in real world images.
They proposed the idea of an inverse detector which combined bottom-up edges and top-down detection.
More recently,~\cite{yu2017casenet} extended the CNN-based class-agnostic edge detector proposed in~\cite{xie2015hed}, and allowed each edge pixel to be associated with more than one class. 
The proposed CASENet architecture combined low and high-level features with  a multi-label loss function to supervise the fused activations. 

Most works use non-maximum-suppression 
\cite{Canny:1986}
as a postprocessing step in order to deal with the thickness of predicted boundaries. In our work, we  directly optimize for NMS during training. We further reason about misaligned ground-truth annotations with real object boundaries, which is typically not done in prior work. Note that our focus here is not to propose a novel edge-detection approach, but rather to have a simple add-on to existing architectures. 

The work most closely related to ours is SEAL~\cite{yu2018seal}, in that it deals with misaligned labels during training.
Similar to us, SEAL treats the underlying ground truth boundaries as a latent variable that is jointly optimized during training.
Optimization is formulated as a computationally expensive bipartite graph min-cost assignment problem. In order to make optimization tractable, there are no pair-wise costs, \ie two neighboring ground-truth pixels can be matched to two pixels far apart in the latent ground-truth, potentially leading to ambiguities in training.
In our work, we infer true object boundaries via a level set formulation which preserves connectivity and proximity, and ensures that the inferred ground-truth boundaries are well behaved. 
Moreover, SEAL is limited to the domain of boundary detection and needs to have reasonably well annotated data, since alignment is defined as a one-to-one mapping between annotated and inferred ground-truth. In our method, substantial differences (topology and deviation) in ground truth can be handled. Our approach can thus be naturally used to refine coarse segmentation labels,  
lending itself as a novel way to efficiently annotate datasets.

{\bf Level Set Segmentation.}
Level Set Methods~\cite{osher1988fronts} have been widely used for image segmentation \cite{caselles1997geodesic,Cremers-11,rupprecht2016deep,hu2017deep,marcos2018learning,Bergtholdt-et-al-06,Li_TIP08,dubrovina2015multi} due to their ability
to automatically handle various topological changes such as splitting and merging. Most older work derived different level set formulations on top of standard image gradient observations, while recent work swapped those with neural network outputs~\cite{hu2017deep}. In~\cite{marcos2018learning}, the authors proposed a deep structured active contours method that learns the parameters of an active contour model using a CNN.~\cite{Geodesic14} introduced a method for object proposal generation, by learning to efficiently place seeds such that critical level sets originating from these seeds hit object boundaries. In parallel work,~\cite{zian19levelset} learns CNN feature extraction and levelset evolution in an end-to-end fashion for object instance annotation. 
In our work, we exploit level set optimization during training as a means to iteratively refine ground-truth semantic boundaries. 

%
%
%
%
%
%
%

%
%
%

%


\begin{figure*}[t!]
\vspace{-2mm}
\centering
\includegraphics[width=0.86\linewidth]{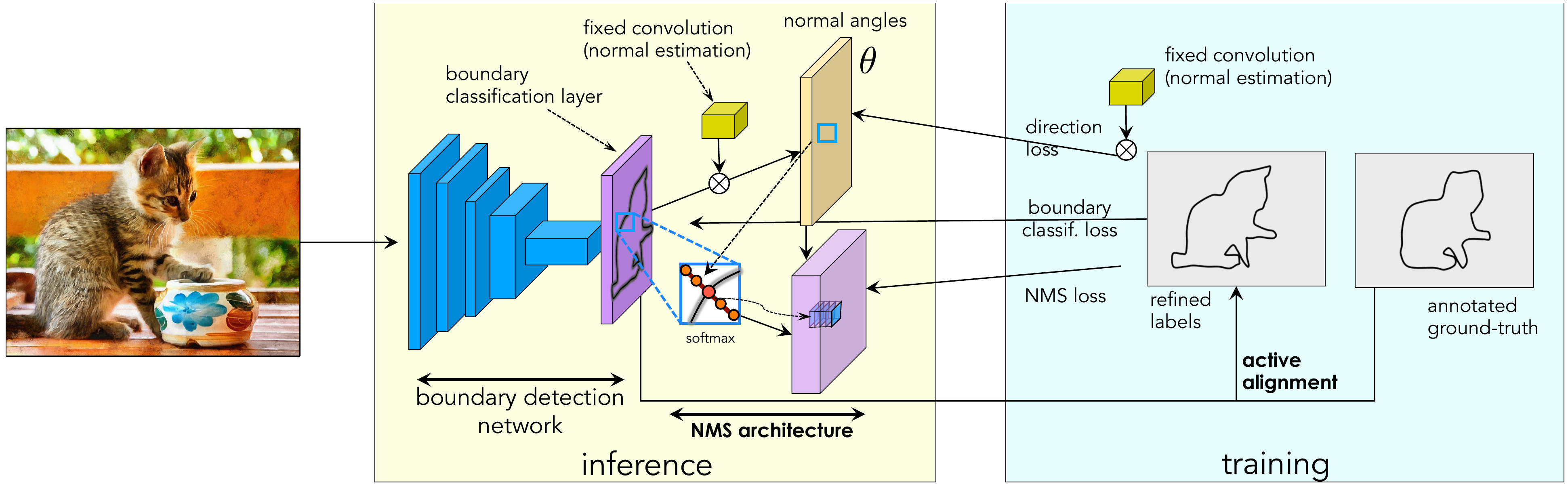}
\vspace{-3mm}
\caption{{\bf STEAL architecture.} Our architecture plugs on top of any backbone architecture. The boundary thinning layer acts upon boundary classification predictions by computing the edge normals, and sampling 5 locations along the normal at each boundary pixel. We perform softmax across these locations, helping us enhance the boundary pixels as in standard NMS. During training, we iteratively refine ground-truth labels using our predictions via an active alignment scheme. NMS and normal direction losses are applied only on the (refined) ground-truth boundary locations.}
\label{fig:nms_layer}
\vspace{-2mm}
\end{figure*}

\vspace{-1mm}
\section{The STEAL Approach}
In this section, we introduce our Semantically Thinned Edge Alignment Learning (STEAL) approach.
Our method consists of a new boundary thinning layer together with a loss function that aims to produce thin and precise semantic edges. We also propose a framework that jointly learns object edges while learning to align noisy human-annotated edges with the true boundaries during training. We refer to the latter as \emph{active alignment}. Intuitively, by using the true boundary signal to train the boundary network, we expect it to learn and produce more accurate predictions. STEAL is agnostic to the backbone CNN architecture, and can be plugged on top of any existing learning-based boundary detection network. We illustrate the framework in Fig.~\ref{fig:nms_layer}. 

Subsec.~\ref{sec:sbd} gives an overview  of semantic boundary detection and the relevant notation. Our boundary thinning layer and loss are introduced  in Subsec.~\ref{sec:nms}. In Subsec.~\ref{sec:alignment}, we describe our active alignment framework.

\vspace{-1mm}
\subsection{Semantic Aware Edge-Detection}
\label{sec:sbd}

Semantic Aware Edge-Detection~\cite{yu2017casenet,yu2018seal} can be defined as the task of predicting boundary maps for $K$ object classes given an input image $\mathbf{x}$. Let $y_k^m\in\{0,1\}$ indicate whether pixel $m$ belongs to class $k$. We aim to compute the probability map $P(\mathbf{y}_k|\mathbf{x}; \mathbf{\theta})$, which is typically assumed to decompose into a set of pixel-wise probabilities $P(y_k^m|\mathbf{x}; \mathbf{\theta})$ modeled by Bernoulli distributions. It is computed with a convolutional neural network $f$ with $k$ sigmoid outputs, and parameters $\mathbf{\theta}$. Each pixel is thus allowed to belong to multiple classes, dealing with the cases of multi-class occlusion boundaries. 
Note that the standard class-agnostic edge detection can be seen as a special case with $k=1$ (consuming all foreground classes). 

\vspace{-3.5mm}
\paragraph{Semantic Edge Learning.}
State-of-the-art boundary detectors are typically trained using the standard binary cross entropy loss adopted from HED~\cite{xie2015hed}.
To deal with the high imbalance between the edge and non-edge pixels, a weighting term $\beta=|Y^{-}|/|Y|$ is often used, where $|Y^{-}|$ accounts for the number of non-edge pixels among all classes in the mini-batch, and $|Y|$ is the total number of pixels.
In the multi-class scenario, the classes are assumed to be independent~\cite{yu2017casenet,yu2018seal}.
Therefore, in learning the following weighted binary cross-entropy loss is minimized:
\vspace{-1mm}
\begin{equation}\label{mle_edges}
\begin{split}
\mathcal{L}_{BCE}(\mathbf{\theta}) &= 
 -\sum_k\log P(\mathbf{y}_k  |\mathbf{x}; \mathbf{\theta}) \\
										 &=  -\sum_{k}\sum_{m} \{ \beta\, y_{k}^m\log f_k(m|\mathbf{x},\mathbf\theta)+ \\
	  									& + (1-\beta) (1-y_{k}^m)\log(1-f_k(m|\mathbf{x},\mathbf\theta))\}
\end{split}
\end{equation} 
where $\mathbf{y}$ indicates the ground-truth boundary labels. 
%
\subsection{ Semantic Boundary Thinning Layer}

In the standard formulation, nearby pixels in each boundary map are considered to be independent, and can cause the predictions to ``fire'' densely around object boundaries.
 We aim to encourage predictions along each boundary pixel's normal to give the maximal response on the actual boundary. This is inspired by edge-based non-maximum suppression (NMS) dating back to Canny's work~\cite{Canny:1986}. Furthermore, we add an additional loss term that encourages the normals estimated from the predicted boundary maps to agree with the normals computed from ground-truth edges. The two losses work together in producing sharper predictions along both the normal and tangent directions.

\subsection{Thinning Layer and NMS Loss}
\label{sec:nms}

	Formally, during training we add a new deterministic layer on top of the boundary prediction map. 
	For each positive ground-truth boundary pixel $p$
	for class $k$ we normalize the responses along the normal direction $\vec{d}_{p}^k$ as follows:	
	%
 	%
 	 \begin{align}
 	h_k(p|\mathbf{x},\theta)= \frac{ \exp (f_k(p|\mathbf{x},\theta)/\tau)}{\sum_{t=-L}^{L} \exp (f_k(p_t|\mathbf{x},\theta)/\tau)}
 	\label{eq:nms}
\end{align}
where:
\vspace{-3mm}
\begin{align}
		x(p_t)&=x(p) + t\cdot \cos \vec{d}_{p}  \\
		y(p_t)&=y(p) + t\cdot \sin \vec{d}_{p}
\end{align}
Here, $t\in\{-L,-L+1,\dots,L\}$, and $L$ denotes the maximum distance of a pixel $p_t$ from $p$ along the normal.  See Fig.~\ref{fig:nms_layer} for a visualization. We compute the normal direction $\vec{d}_{p}^k$ from the ground-truth boundary map using basic trigonometry and a fixed convolutional layer that estimates second derivatives.
The parameter $\tau$ in Eq.~\eqref{eq:nms} denotes the temperature of the softmax.  We use $L=2$ and $\tau=0.1$. 
Intuitively, we want to encourage the true boundary pixel $p$ to achieve the highest response along its normal direction. 
We do this via an additional loss, referred to as the NMS loss, that pushes the predicted categorical distribution computed with $h$ towards a Dirac delta target distribution: %
\begin{equation}
	\mathcal{L}_{nms}(\theta)= -\sum_k\sum_{p} \log h_k(p|\mathbf{x},\theta)
	\label{nms_final}
\end{equation}
Note that $p$ indexes only the positive boundary pixels for each class, other pixels do not incur the NMS loss. 
We compute $f_k(p_t|\mathbf{x},\theta)$ in Eq.~\eqref{eq:nms} for non-integral locations using a bilinear kernel. 

\vspace{-3mm}
\paragraph{Direction Loss.} Ideally, the predicted boundaries would have normal directions similar to those computed from the ground-truth boundaries. We follow~\cite{DWT17} to define the error as the mean squared loss function in the angular domain:
\begin{align}
\mathcal{L}_{\mathrm{dir}}(\theta)&=\sum_k\sum_p ||\cos^{-1} \langle\vec{d_p},\vec{{e}_p}(\theta)\rangle||, 
\end{align}
with $\vec{d_p}$ the ground-truth normal direction in boundary pixel $p$, and $\vec{e_p}$ the normal computed from the predicted boundary map. We use the same convolutional layer on top of $f_k$ to get $\vec{e}$. 
Finally, we compute our full augmented loss as the combination of the following three terms:
\begin{align}\label{eq_all_losses}
\mathcal{L}= \alpha_1\,\mathcal{L}_{BCE} + \alpha_2\,\mathcal{L}_\textnormal{nms} +\alpha_3\,\mathcal{L}_\textnormal{dir}
\end{align}
where $\alpha_1,\alpha_2,\alpha_3$ are hyper-parameters that control the importance of each term (see Experiments).

\subsection{Active Alignment}
\label{sec:alignment}

Learning good boundary detectors requires high quality annotated data. 
However, accurate boundaries are time consuming to annotate. Thus datasets tradeoff between quality and annotation efficiency. Like~\cite{yu2018seal}, we notice that the standard SBD benchmark~\cite{BharathICCV2011} contains significant label noise. 
In this section, we propose a framework that allows us to jointly reason about true semantic boundaries and train a network to predict them. We adopt a level set formulation which ensures that the inferred ``true'' boundaries remain connected, and are generally well behaved. 

Let  $\byhat= \{\hat{\mathbf{y}}_1,\hat{\mathbf{y}}_2,...,\hat{\mathbf{y}}_K \} $ denote a more accurate version of the ground-truth label $\by$, which we aim to infer as part of our training procedure. We define $\hat{\mathbf{y}}_i$ as a curve.
%
Our goal is to jointly optimize for the latent variable $\byhat$ and parameters $\theta$ of the boundary detection network. 
The optimization is defined as a minimization of the  following loss:
\vspace{-2.3mm}
\begin{align}\label{logpmin}
\min_{\hat{\mathbf{y}}, \mathbf{\theta}}\mathcal{L}(\hat{\mathbf{y}},\mathbf{\theta}) &= - \sum_k \log P(\mathbf{y}_k, \hat{\mathbf{y}}_k|\mathbf{x}; \mathbf{\theta}) \\
&= -\sum_k \big(\log P(\mathbf{y}_k|\hat{\mathbf{y}}_k) + \log P(\hat{\mathbf{y}}_k|\mathbf{x}; \mathbf{\theta})  \big)\nonumber
\vspace{-2.3mm}
\end{align}

\vspace{-2mm}
The second term is the log-likelihood of the model and can be defined as in the previous section.
The first term encodes the prior that encourages $\hat{\mathbf{y}}_k$ to be close to ${\mathbf{y}}_k$. Inspired by~\cite{caselles1997geodesic}, we define this term with an energy that describes ``well behaved'' curves:
\vspace{-1.5mm}
\begin{equation}\label{energy_prior}
	E(\mathbf{y}_k|\hat{\mathbf{y}}_k,\lambda) = \int_{q} g_k(\hat{\mathbf{y}}_k(q)) \ |\hat{\mathbf{y}}'_k(q)| \ \partial{q}
\vspace{-2.5mm}
\end{equation}
where we define $g_k(.)$ as the following decreasing function:
\vspace{-2mm}
\begin{equation}\label{g_function}
	g_k= \frac{1}{\sqrt{1+f_k}} + \frac{\lambda}{\sqrt{1+\mathbf{y}_k}}
\end{equation}
Here, $\lambda$ is a hyper-parameter that controls the effect of $\mathbf{y}_k$. 
Intuitively, this energy is minimized when the curve $\hat{\mathbf{y}}_k$ lies in areas of high probability mass of $f_k$,
 and is close, by a factor of $\lambda$, to the given ground-truth $\mathbf{y}_k$. 

We can minimize Eq.~\eqref{energy_prior} via steepest-descent, where
we find the gradient descent direction that allows to deform the initial $\hat{\mathbf{y}}^0_k$ (here we use the given noisy ground-truth as the initial curve) towards a (local) minima of Eq.~\eqref{energy_prior}~\cite{caselles1997geodesic}:
\vspace{-1.5mm}
\begin{equation}
\label{eq:levelset}
\frac{\partial{\hat{\mathbf{y}}_k^t}} {\partial{t}}=  g_k\, \kappa\ \vec{\mathbf{n}} - (\nabla g_k \cdot \vec{\mathbf n})\vec{\mathbf n}
\vspace{-1.5mm}
\end{equation}
Here $\kappa$ is the Euclidean curvature and $\vec{\mathbf{n}}$ is the inward normal to the boundary. Details of this computation can be found in~\cite{caselles1997geodesic}, Appendix B and C.

Eq~\eqref{eq:levelset} follows the level-set approach~\cite{osher1988fronts}, where the curve $\hat{\mathbf{y}}^t_k$ is a $0$ level-set of an embedding function $\phi$, \ie a set of points satisfying $\phi(.)=0$. By differentiating the latter equation, it is easy to show that if $\hat{\mathbf{y}}_k$  evolves according to $\frac{\partial{\hat{\mathbf{y}}_k^t}} {\partial{t}}=\beta \vec{\mathbf {n}}\,$ then the embedding function $\phi$ can be deformed as $\frac{\partial{{\phi}}} {\partial{t}} = \beta \vec{|\nabla \phi|} $~\cite{caselles1997geodesic}. 
We can thus rewrite  the evolution of $\hat{\mathbf{y}}_k$ \ in terms of $\phi$ as follows:
\vspace{-1mm}
\begin{equation}\label{level_set_equation}
%
%
%
\frac{\partial{\phi}} {\partial{t}}= g_k( \kappa+c) |\nabla \phi|  + \nabla g_k \cdot \nabla \phi
\vspace{-1mm}
\end{equation}
where $c$ can be seen as a constant velocity that helps to avoid certain local minima~\cite{caselles1997geodesic}. Eq.~\ref{level_set_equation} can also be interpreted as the Geodesic Active Contour formulation of the Level Set Method~\cite{caselles1997geodesic,osher1988fronts}.
\subsection{Learning}
Minimizing Eq.~\eqref{logpmin} can be performed with an iterative two step optimization process.
In one step, we 
evolve the provided boundary $\mathbf{y}_k$ towards areas where the network is highly confident. The number of evolution steps indexed by $t$ can be treated as a latent variable and $\hat{\mathbf{y}}_k$ is selected by choosing the $\byhat_k^t$ that minimizes Eq.~\eqref{logpmin}. 
In the second step, we optimize $\mathbf{\theta}$  using  the computed ${\mathbf{y}_k}$. 

Formally, we want to solve:
\vspace{-1mm}
\begin{equation}\label{mineq1}
\begin{split}
\min_{\hat{\mathbf{y} }, \mathbf{\theta}}\mathcal{L}(\hat{\mathbf{y}} ,\mathbf{\theta}) &=
\min_{ \mathbf{\theta}}\min_{\hat{\mathbf{y}} }\mathcal{L}(\hat{\mathbf{y}} ,\mathbf{\theta})
\end{split}
\end{equation}
where we iterate between holding $\theta$ fixed and optimizing $\hat{\mathbf{y}}$:
\begin{equation}\label{mineq2}
\begin{split}
\min_{\hat{\mathbf{y}}_k}\mathcal{L}(\hat{\mathbf{y}}_k,\mathbf{\theta}) &= \min_{t}  \{-\log P(\hat{\mathbf{y}}_{k}^t|\mathbf{x}; \mathbf{\theta}) -C  \} 
%
%
%
\end{split}
\end{equation}
and optimizing $\theta$ via Eq.~\eqref{eq_all_losses} while holding $\hat{\mathbf{y}}$ fixed. 
Here $C$ is a constant that does not affect optimization. 
%

%
\subsection{Coarse-to-Fine Annotation}
\label{sec:coarse2fine}
Embedding the evolution of $\byhat$ \ in that of $\phi$
has two main benefits.
Firstly, topological changes of $\byhat$ are handled for free and accuracy and stability can be achieved by using proper numerical methods. Secondly, $\phi$ can be naturally interpreted as a mask segmenting an object, where $\phi<0$ corresponds to the segmented region. Moreover, our approach can also be easily used to speed up object annotation. Assume a scenario where an annotator draws a coarse mask inside an object of interest, by using only a few clicks. This is how the coarse subset of the Cityscapes dataset has been annotated~\cite{cityscapes}. We can use our learned model and levelset formulation (Eq.~\eqref{level_set_equation}), setting $\lambda=0$ and $c=1$ to evolve the given coarse mask by $t$ iterations to produce an improved segmentation mask whose edges align with the edges predicted by our model.

\begin{table*}[t!]
\vspace{-2.5mm}
\resizebox{\linewidth}{!}{
\addtolength{\tabcolsep}{-2.2pt}
\begin{tabular}{c|c|c|c|c|c|c|c|c|c|c|c|c|c|c|c|c|c|c|c|c|c|c}
Metric & Method &   aero &   bike &   bird &   boat &  bottle &    bus &    car &    cat &  chair &    cow &  table &    dog &  horse &  mbike &  person &  plant &  sheep &   sofa &  train &     tv &   mean \\
\hline \hline

\multirow{4}{0.07\linewidth}{\centering{MF\\(ODS)}}  
  & CASENet & 74.84   & 60.17   & 73.71   & 47.68   & 66.69   & 78.59   & 66.66   & 76.23   & 47.17   & 69.35   & 36.23   & 75.88   & 72.45   & 61.78   & 73.10   & 43.01   & 71.23   & 48.82   & 71.87   & 54.93   & 63.52 \\ 
  & CASENet-S & 76.26   & 62.88   & 75.77   & 51.66   & 66.73   & 79.78   & 70.32   & 78.90   & 49.72   & 69.55   & 39.84   & 77.25   & 74.29   & 65.39   & 75.35   & 47.85   & 72.03   & 51.39   & 73.13   & 57.35   & 65.77 \\ 
  & SEAL & 78.41   & 66.32   & 76.83   & 52.18   & 67.52   & 79.93   & 69.71   & \bf{79.37}   & 49.45   & 72.52   & 41.38   & 78.12   & 74.57   & 65.98   & 76.47   & 49.98   & 72.78   & \bf{52.10}   & 74.05   & 58.16   & 66.79 \\ 

\cline{2-23}
  & Ours (NMS Loss) & 78.96   & 66.20   & 77.53   & \bf{54.76}   & 69.42   & \bf{81.77}   & \bf{71.38}   & 78.28   & \bf{52.01}   & \bf{74.10}   & 42.79   & 79.18   & 76.57   & 66.71   & 77.71   & 49.70   & 74.99   & 50.54   & \bf{75.50}   & 59.32   & 67.87 \\ 
  & Ours (NMS Loss + AAlign) & \bf{80.15}   & \bf{67.80}   & \bf{77.69}   & 54.26   & \bf{69.54}   & 81.48   & 71.34   & {78.97}   & {51.76}   & 73.61   & \bf{42.82}   & \bf{79.80}   & \bf{76.44}   & \bf{67.68}   & \bf{78.16}   & \bf{50.43}   & \bf{75.06}   &  {50.99}   & 75.31   & \bf{59.66}   & \bf{68.15} \\ 
\hline \hline

\multirow{4}{0.07\linewidth}{\centering{AP}}
  & CASENet & 50.53   & 44.88   & 41.69   & 28.92   & 42.97   & 54.46   & 47.39   & 58.28   & 35.53   & 45.61   & 25.22   & 56.39   & 48.45   & 42.79   & 55.38   & 27.31   & 48.69   & 39.88   & 45.05   & 34.77   & 43.71 \\   
  & CASENet-S & 67.64   & 53.10   & 69.79   & 40.51   & 62.52   & 73.49   & 63.10   & 75.26   & 39.96   & 60.74   & 30.43   & 72.28   & 65.15   & 56.57   & 70.80   & 33.91   & 61.92   & 45.09   & 67.87   & 48.93   & 57.95 \\ 
  & SEAL & 74.24   & 57.45   & 72.72   & 42.52   & 65.39   & 74.50   & 65.52   & \bf{77.93}   & 40.92   & 65.76   & \bf{33.36}   & 76.31   & 68.85   & 58.31   & 73.76   & 38.87   & 66.31   & \bf{46.93}   & 69.40   & \bf{51.40}   & 61.02 \\ 
  \cline{2-23}
  & Ours (NMS Loss) & 75.85   & 59.65   & \bf{74.29}   & \bf{43.68}   & 65.65   & \bf{77.63}   & 67.22   & 76.63   & \bf{42.33}   & \bf{70.67}   & 31.23   & 77.66   & 74.59   & 61.04   & 77.44   & 38.28   & 69.53   & 40.84   & \bf{71.69}   & 50.39   & 62.32 \\ 
  & Ours (NMS Loss + AAlign) & \bf{76.74}   & \bf{60.94}   & 73.92   & 43.13   & \bf{66.48}   & 77.09   & \bf{67.80}   & 77.50   & 42.09   & 70.05   & 32.11   & \bf{78.42}   & \bf{74.77}   & \bf{61.28}   & \bf{77.52}   & \bf{39.02}   & \bf{68.51}   & 41.46   & 71.62   & 51.04   & \bf{62.57} \\ 
\hline \hline

\end{tabular}
}
\centering
\vspace{-3mm}
\caption{Comparison of our method in the re-annotated SBD test set vs state-of-the-art. Scores are measured by $\%$.\label{tb:main}}
\end{table*}

\begin{table*}[t!]
\vspace{-2mm}
\centering
\resizebox{\textwidth}{!}{\begin{tabular}{c|c|c|c|c|c|c|c|c|c|c|c|c|c|c|c|c|c|c|c|c|c}
Method & aero & bike & bird & boat & bottle & bus & car & cat & chair & cow & table & dog & horse & mbike & person & plant & sheep & sofa & train & tv & mean\\
\hline \hline
CASENet \cite{yu2017casenet} & 83.3 & 76.0 & 80.7 & 63.4 & 69.2 & 81.3 & 74.9 & 83.2 & 54.3 & 74.8 & 46.4 & 80.3 & 80.2 & 76.6 & 80.8 & 53.3 & 77.2 & 50.1 & 75.9 & 66.8 & 71.4\\
SEAL \cite{yu2018seal} & {84.9} &  {78.6} &  {84.6} &  {66.2} &  {71.3} & {83.0} &  {76.5} &  {87.2} &  {57.6} &  {77.5} &  {53.0} &  {83.5} &  {82.2} &  {78.3} &  {85.1} &  {58.7} &  {78.9} & {53.1} &  {77.7} &  {69.7} & {74.4}\\
\hline
Ours & \textbf{85.8} & \textbf{80.0} & \textbf{85.6} & \textbf{68.4} & \textbf{71.6} & \textbf{85.7} & \textbf{78.1} & \textbf{87.5} & \textbf{59.1} & \textbf{78.5} & \textbf{53.7} & \textbf{84.8} & \textbf{83.4} & \textbf{79.5} & \textbf{85.3} & \textbf{60.2} & \textbf{79.6} & \textbf{53.7} & \textbf{80.3} & \textbf{71.4} &  \textbf{75.6}

\end{tabular}
}
\vspace{-3mm}
\caption{Results on SBD test following the original evaluation protocol, and test set from~\cite{BharathICCV2011}.\label{sbd_sota}}
\vspace{-2mm}
\end{table*}

\vspace{-2mm}
\subsubsection{Implementation Details}
\paragraph{Morphological Level Set.} In this work, we follow a morphological approach to compute the differential operators used in the curve's evolution. 
This solution is based on numerical methods which are simple, fast and stable. 
Additionally, in this approach, the level set is just a binary piecewise constant function and constant reinitialization of the level set function is not required. 
We refer the reader to~\cite{marquez2014morphological} for a more detailed explanation and implementation details.

\vspace{-4mm}
\paragraph{Training Strategy.} 
Our active alignment heavily relies on the quality of the network's predictions to iteratively refine the noisy ground-truth. 
During initial stages of training, the network is not confident  and may lead us to  infer potentially noisier labels.
We hence introduce alignment  after the network's accuracy starts to flatten.
In our formulation, this can be seen as setting $\lambda=\inf$ for a certain number of iterations.
In order to  save on computation time, active alignment can also be applied every $n$ training iterations.

%

\vspace{-1mm}
\section{Experimental Results}
\vspace{-1mm}
%
%
%
%
In this section, we provide an extensive evaluation of our approach on the standard SBD benchmark~\cite{BharathICCV2011}, as well as on the Cityscapes dataset~\cite{cityscapes}.
We further show how our approach can be  used to significantly improve coarse segmentation labels, mimicking a scenario where we train on a labeled dataset with moderate noise, and use the trained model to generate finer annotations from only coarsely annotated data (collected with  less manual annotation effort). 

\begin{table}
\vspace{-4mm}
\begin{minipage}{0.45\textwidth}
\resizebox{\linewidth}{!}{
\begin{tabular}{c|c|c|c|c}
Metric & Method & Test NMS & Or. Test Set   &  Re-annot. Test Set \\
\hline \hline
\multirow{4}{0.07\linewidth}{\centering{ MF\\(ODS)}}  
 & CASENet &  & 62.21 & 63.52\\
\cline{2-5}
 & Ours (CASENet) &  & 63.20 & 64.03 \\
 & Ours (CASENet) & \checkmark & 64.84 &66.58\\
 \cline{2-5}
 & + NMS Layer &  & 64.15 & 64.99 \\
 & + NMS Layer & \checkmark &  \textbf{65.93} & {67.87}  \\
\cline{2-5}
 & + Active Align & \checkmark & 64.83  & \bf{68.15}  \\
\hline \hline

\multirow{4}{0.07\linewidth}{\centering{AP}}
   & CASENet &  & 42.99 & 43.71 \\
   \cline{2-5} 
   & Ours (CASENet) &  & 34.60  & 45.60 \\
   & Ours (CASENet) & \checkmark & 44.83   & 60.48 \\

 \cline{2-5}
 & + NMS Layer &  & 53.67  & 54.18\\
 & + NMS Layer & \checkmark &  \textbf{60.10} & {62.32}\\
 \cline{2-5}
 & + Active Align & \checkmark & 57.98  & \bf{62.57}\\

\hline \hline

\end{tabular}
}
\centering
\vspace{-3mm}
\caption{Effect of the NMS Loss and Active Alignment on the SBD dataset. Score (\%)  represents mean over all classes.}
\label{tbl_effect_nms_loss_and_alignment}
\end{minipage}
\vspace{-2mm}
\end{table}

\vspace{-3.5mm}
\paragraph{Implementation Details.} In all experiments, we select CASENet~\cite{yu2017casenet} as the backbone network  since it is the current  state-of-the-art semantic-aware-edge detection architecture.
We re-implement CASENet in PyTorch following~\cite{yu2017casenet}.
The performance of our reimplementation (slightly better) is illustrated in tables as \textit{CASENet Ours} for fair comparison.
We use $472\times472$ as the training resolution. 
Training is done on an NVIDIA DGX Station using 4 GPUs with a total batch size of 8. We use $\alpha_1=1, \alpha_2=10,\alpha_3=1$ in our loss function.
For SBD, we use a learning rate of 1e-7. %
At 20k iter, we decrease the learning rate by a factor of 10 and set $\beta=0$. 
Active alignment is done  every $5$k iter ($\lambda=1$) starting at 55k iter.
The full model converges at about $70k$ iter and takes approximately two days to train.
For Cityscapes, we set the learning rate to be  5e-8, and decay is done every $20k$ iterations  by a factor of 20. 
Since images are more densely annotated, we set the weights of the loss function to be 1.
We do not use active alignment in Cityscapes since the train set is finely annotated.
This is used later for the refinement of coarse data.
The model converges at around $60$k iterations.

\begin{table}
\vspace{-4mm}
\begin{minipage}{0.45\textwidth}
\resizebox{\linewidth}{!}{
\begin{tabular}{c|c|c|c|c }
Metric & Method & Active Align & Noisy Train &  Noisy Train ( +8px err)   \\
\hline \hline
\multirow{3}{0.1\linewidth}{\centering{MF\\(ODS)}} 
 & Ours (CASENet) &  & 64.03  & 50.58    \\
 & Ours (CASENet) & \checkmark  & 64.10  & 52.69  \\
 & + NMS Layer &  \checkmark  & \bf{68.15} &  \bf{56.41} \\
 
\hline \hline

\multirow{3}{0.1\linewidth}{\centering{AP}}
   & Ours (CASENet) &  & 45.60   & 29.32     \\
   & Ours (CASENet) & \checkmark & 45.41     & 27.60   \\

 & + NMS Layer & \checkmark & \bf{62.57}   &  \bf{43.97} \\

\hline \hline

\end{tabular}
}
\centering
\vspace{-3.5mm}
\caption{Effect of Active Alignment on the SBD dataset. Score (\%)  represents mean over all classes.}
\label{tbl_effect_active_alignment}
\end{minipage}
\vspace{-3.5mm}
\end{table}

\begin{figure*}[t!]
\vspace{-3mm}
\centering
\includegraphics[width=\linewidth,height=1.9cm,trim=0 0 0 20,clip]{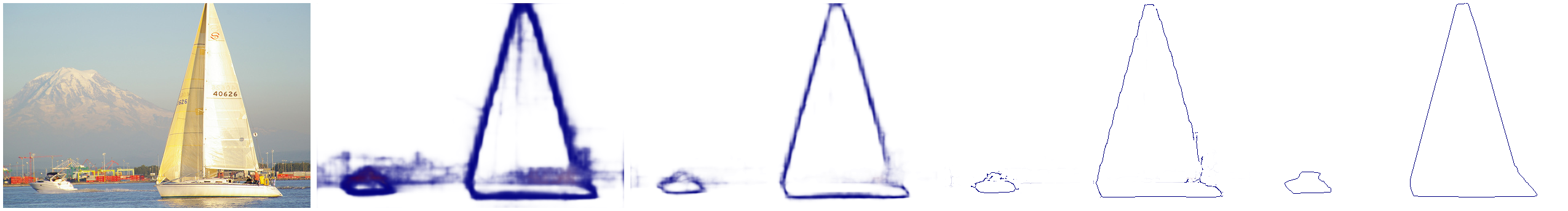}
\includegraphics[width=\linewidth,height=1.9cm]{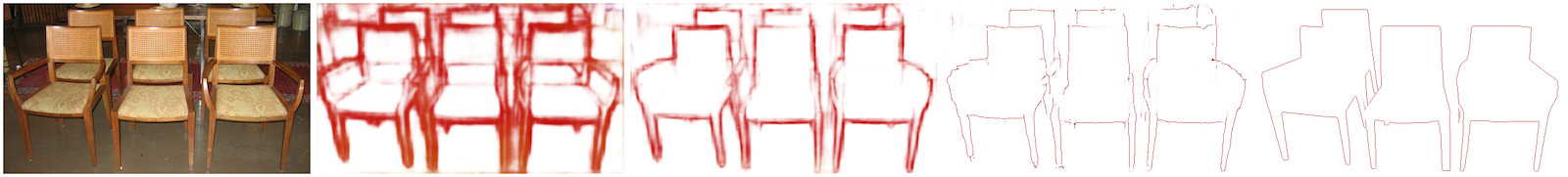}
\includegraphics[width=\linewidth,height=2.2cm]{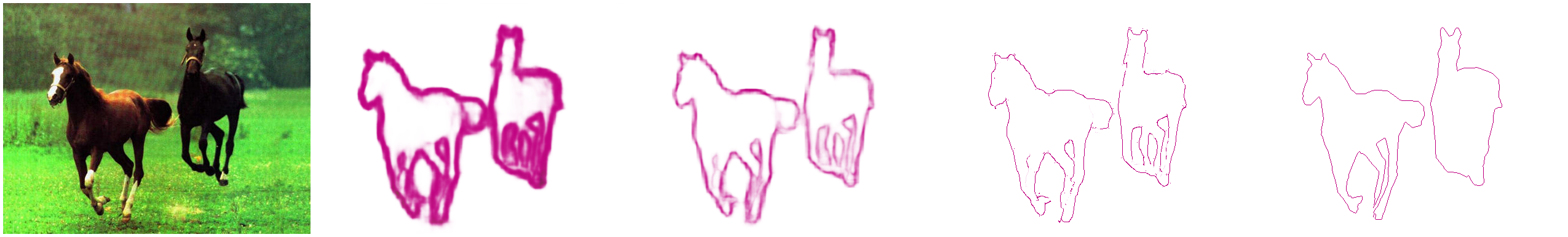}
\includegraphics[width=\linewidth,trim=0 50 0 50,clip]{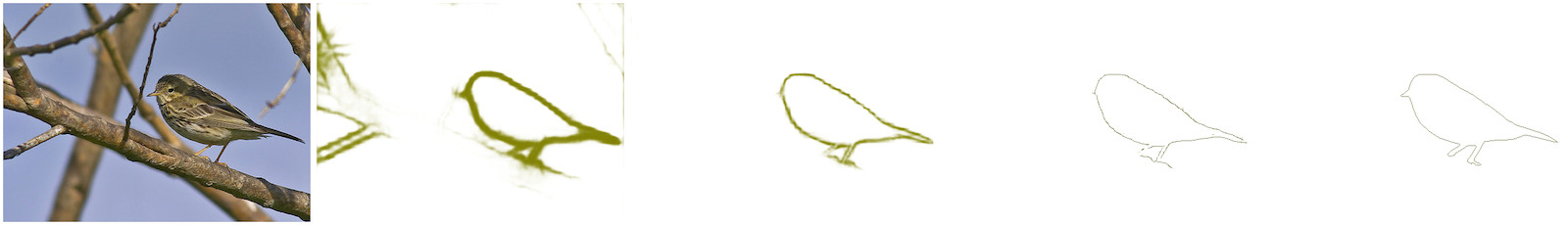}
\begin{small}
\vspace{-2mm}
\centering
\hspace{0.8cm}\begin{tabular}{p{3.0cm}p{3.2cm}p{2.5cm}p{3.2cm}p{3.8cm}}
(a) Image & (b) CASENet & (c) Ours & (d) +Thinning Layer & (e) Ground-truth
\end{tabular}
\end{small}
\vspace{-5mm}
\caption{Qualitative Results on the SBD Dataset.}
\label{fig:sbd_img2}
\end{figure*}  %

\begin{figure*}[t!]
\vspace{-3mm}
\addtolength{\tabcolsep}{-4.2pt}
\begin{tabular}{cccccc}
\includegraphics[height=2.0cm, trim=105 240 50 280,clip]{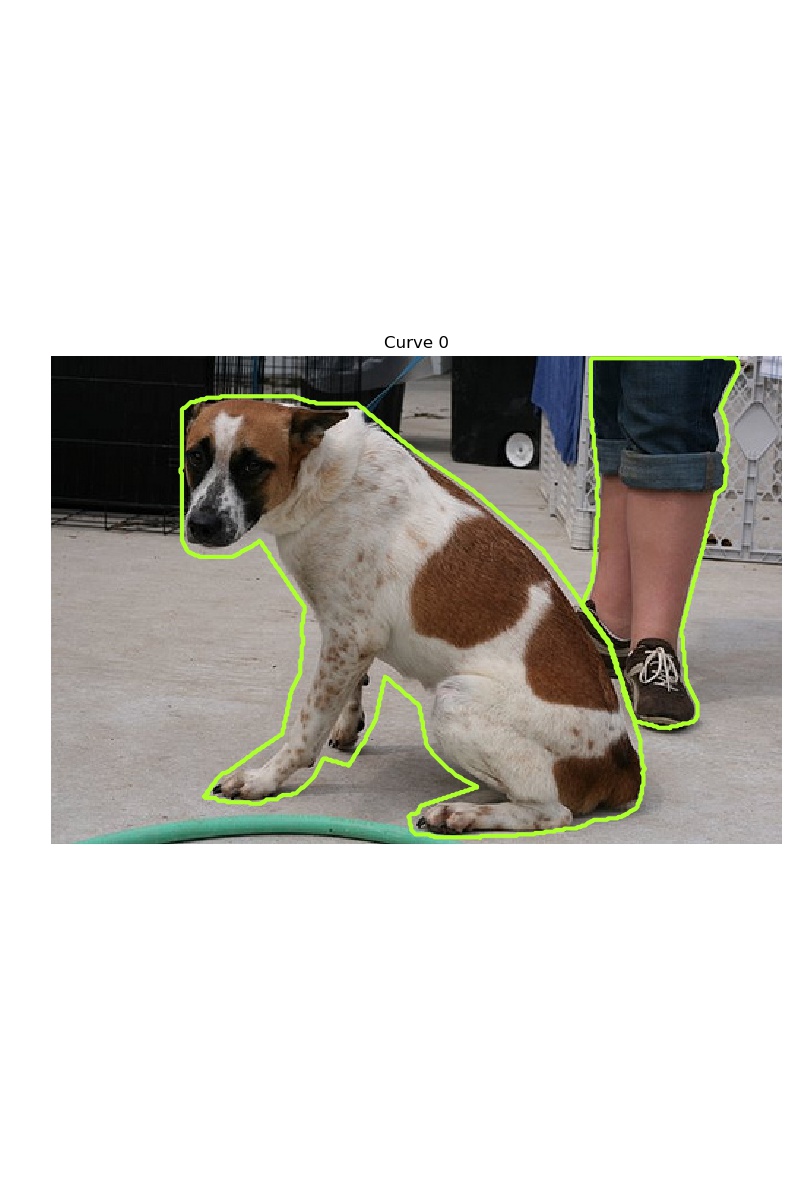}&
\includegraphics[height=2.0cm, trim=105 240 50 280,clip]{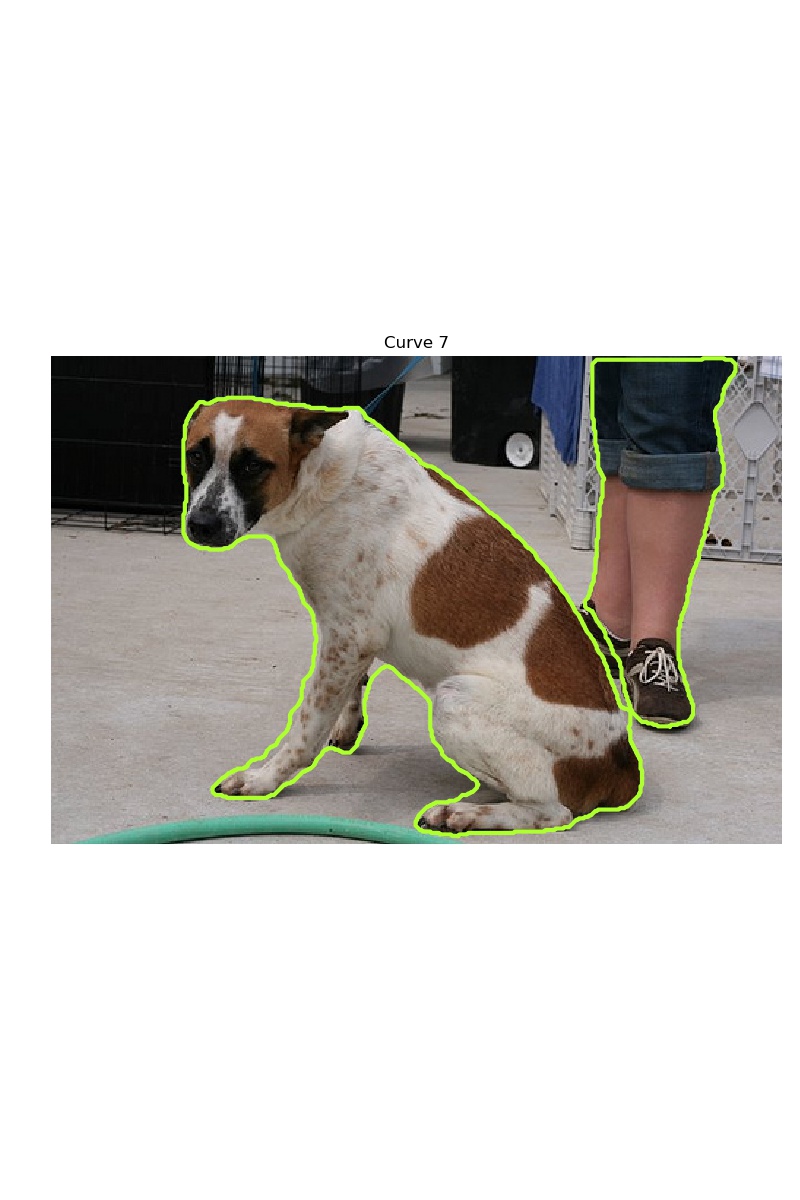}&
\includegraphics[height=2cm, trim=165 580 115 40,clip]{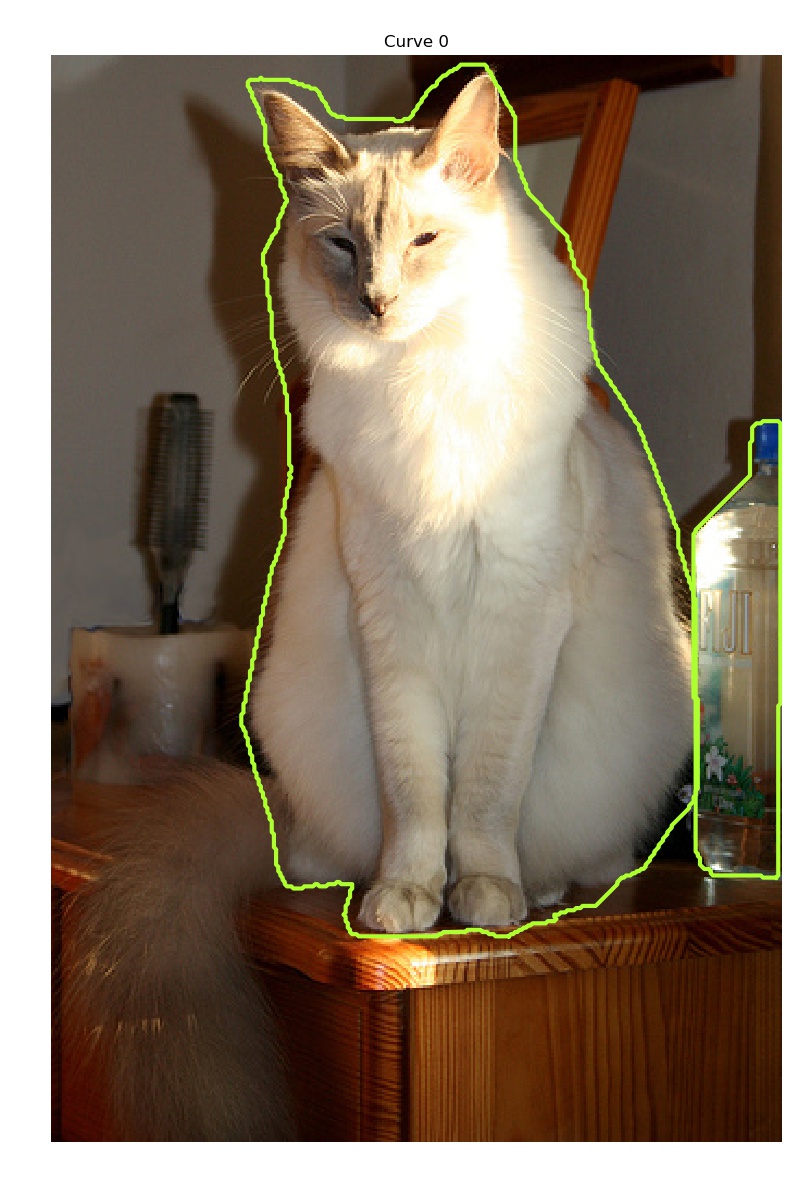}&
\includegraphics[height=2cm, trim=165 580 115 40,clip]{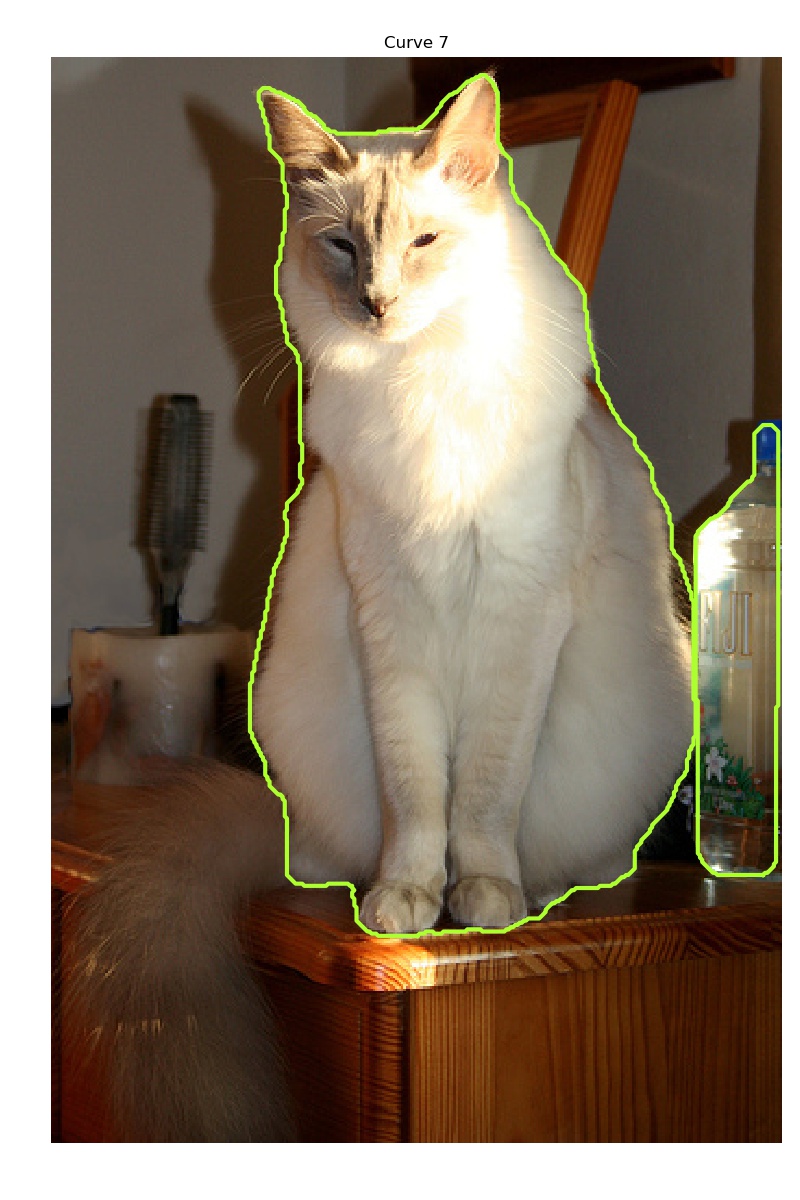}&
\includegraphics[height=2cm, trim=60 330 80 280,clip]{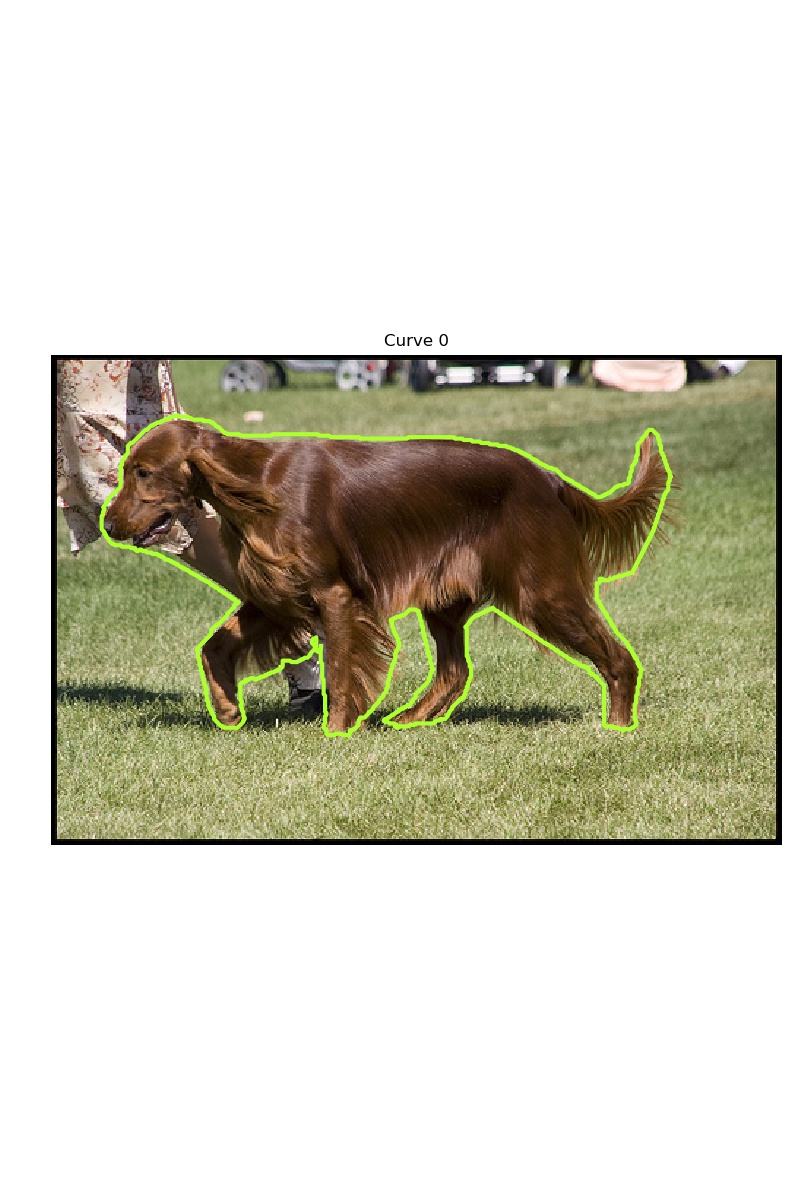}&
\includegraphics[height=2cm, trim=60 330 80 280,clip]{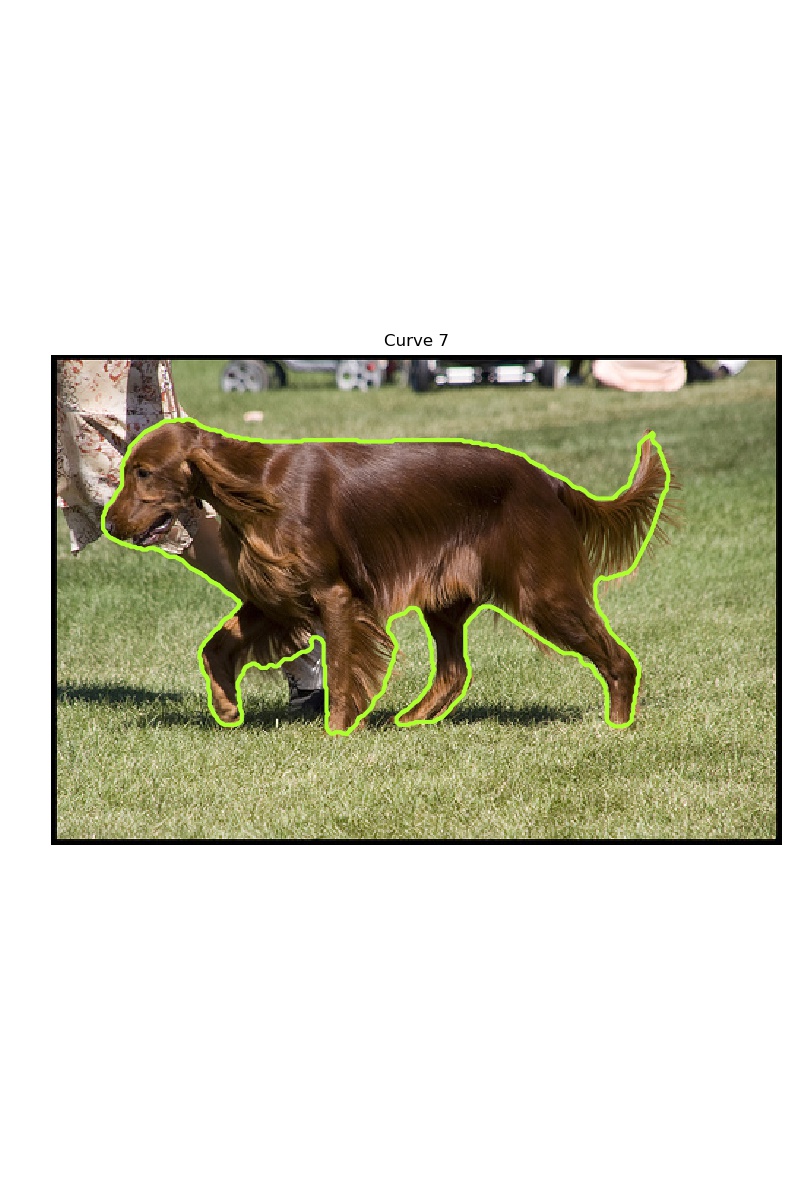}
\end{tabular}

\vspace{-4mm}
\caption{Active Alignment. From Left-to-right (GT, Refined).}
\label{fig:sbd_active_align}
\vspace{-3mm}
\end{figure*}

\begin{table*}[t!]
\centering
\addtolength{\tabcolsep}{-2pt}
\resizebox{\textwidth}{!}{\begin{tabular}{c|l|c|c|c|c|c|c|c|c|c|c|c|c|c|c|c|c|c|c|c|c|c}
Metric & Method & Test NMS & road & s.walk & build. & wall & fence & pole & t-light & t-sign & veg & terrain & sky & person & rider & car & truck & bus & train & motor & bike & mean \\
\hline \hline
\multirow{3}{0.1\linewidth}{\centering{MF\\(ODS)}}
&        CASENet & & 87.06   & 75.95   & 75.74   & 46.87   & 47.74   & 73.23   & 72.70   & 75.65   & 80.42   & 57.77   & 86.69   & 81.02   & 67.93   & 89.10   & 45.92   & 68.05   & 49.63   & 54.21   & 73.74   & 68.92 \\ 

&  Ours(CASENet)& & 87.23   & 76.08   & 75.73   & 47.86   & 47.57   & 73.67   & 71.77   & 75.19   & 80.58   & 58.39   & 86.78   & 81.00   & 68.18   & 89.31   & 48.99   & 67.82   & 50.84   & 55.30   & 74.16   & 69.29 \\ 
& Ours(CASENet) & \checkmark & 88.13   & 76.53   & 76.75   & 48.70   & 48.60   & 74.21   & 74.54   & 76.38   & 81.32   & 58.98   & 87.26   & 81.90   & 69.05   & 90.27   & 50.93   & 68.41   & 52.11   & 56.23   & 75.66   & 70.31 \\ 
& + NMS LOSS & & 88.08   & 77.62   & 77.08   & 50.02   & 49.62   & 75.48   & 74.01   & 76.66   & 81.51   & 59.41   & 87.24   & 81.90   & 69.87   & 89.50   & 52.15   & 67.80   & \bf{53.60}   & 55.93   & 75.17   & 70.67 \\ 
& + NMS LOSS & \checkmark & \bf{88.94}   & \bf{78.21}   & \bf{77.75}   & \bf{50.59}   & \bf{50.39}   & \bf{75.54}   & \bf{76.31}   & \bf{77.45}   & \bf{82.28}   & \bf{60.19}   & \bf{87.99}   & \bf{82.48}   & \bf{70.18}   & \bf{90.40}   & \bf{53.31}   & \bf{68.50}   & {53.39}   & \bf{56.99}   & \bf{76.14}   & \bf{71.42} \\ 

\hline \hline
\multirow{3}{0.1\linewidth}{\centering{AP}}
&        CASENet & & 54.58   & 65.44   & 67.75   & 37.97   & 39.93   & 57.28   & 64.65   & 69.38   & 71.27   & 50.28   & 73.99   & 72.56   & 59.92   & 66.84   & 35.91   & 56.04   & 41.19   & 46.88   & 63.54   & 57.65 \\ 

&  Ours(CASENet) & & 68.38   & 69.61   & 70.28   & 40.00   & 39.26   & 61.74   & 62.74   & 73.02   & 72.77   & 50.91   & 80.72   & 76.06   & 60.49   & 79.43   & 40.86   & 62.27   & 42.87   & 48.84   & 64.42   & 61.30 \\ 
& Ours(CASENet) & \checkmark & 88.83   & 73.94   & 76.86   & 42.06   & 41.75   & 69.81   & 74.50   & 76.98   & 79.67   & 56.48   & \bf{87.73}   & 83.21   & \bf{68.10}   & 91.20   & 44.17   & 66.69   & 44.77   & 52.04   & \bf{75.65}   & 68.13 \\ 
& +NMS LOSS &  & 89.54   & 75.72   & 74.95   & 42.72   & 41.53   & 65.86   & 67.55   & 75.84   & 77.85   & 52.72   & 82.70   & 79.89   & 62.59   & 91.07   & 45.26   & \bf{67.73}   & \bf{47.08}   & 50.91   & 70.78   & 66.44 \\ 
& +NMS LOSS & \checkmark & \bf{90.86}   & \bf{78.94}   & \bf{77.36}   & \bf{43.01}   & \bf{42.33}   & \bf{71.13}   & \bf{75.57}   & \bf{77.60}   & \bf{81.60}   & \bf{56.98}   & {87.30}   & \bf{83.21}   & {66.79}   & \bf{91.59}   & \bf{45.33}   & {66.64}   & {46.25}   & \bf{52.07}   & {74.41}   & \bf{68.89} \\

\end{tabular}
}
\vspace{-3.0mm}
\caption{Results on the val set on the Cityscapes dataset. Training is done using the finely annotated train set. Scores are measured by \%. \label{cityscapes_sota}}
\vspace{-3.5mm}
\end{table*}

\vspace{-1mm}
\subsection{Datasets and Evaluation Metrics}
\label{sec:datasets}
\vspace{-1mm}

\paragraph{Semantic Boundary Dataset (SBD)~\cite{BharathICCV2011}}
 contains 11355 images from the trainval set of PASCAL VOC2011, with 8498 images divided into training, and 2857 as test. 
This dataset contains annotations following the 20-class definitions in PASCAL VOC.
In our experiments, we randomly select 100 images from the training set, which are used as 
our inference.
Training is performed on the remaining 8398 images and evaluation is done on test.
We additionally report performance on the high-quality re-annotated SBD test set from~\cite{yu2018seal}.  This constitutes 1059 images from SBD test.

\vspace{-4mm}
\paragraph{Cityscapes Dataset~\cite{cityscapes}}
contains  5000 finely annotated images divided into 2975 training, 500 inference, and 1525 test images. Since the boundaries are not provided and test is held-out, we follow~\cite{yu2017casenet} to generate the ground truth edges and use the inference images as our test set.

\vspace{-4mm}
\paragraph{Evaluation Protocol:}
We follow the evaluation protocol proposed in~\cite{yu2018seal} which is considerable harder than the one used in~\cite{BharathICCV2011,amfm_pami2011,yu2017casenet}.
An important parameter is the matching distance tolerance which is defined as the maximum slack allowed for boundary predictions to be considered as correct matches to ground-truth. We follow~\cite{yu2018seal} and set it to be 0.0075 for SBD and 0.0035 for Cityscapes.
For further comparisons, in Table~\ref{sbd_sota} we also report 
the performance with the original SBD evaluation protocol~\cite{BharathICCV2011}. 

\vspace{-3.5mm}
\paragraph{Coarse Label Simulation.} In order to quantify the level of annotation noise that our approach can handle, we synthetically coarsen the given labels following the procedure described in~\cite{Zlateski_2018_CVPR}.
This algorithm, inspired by the way that coarse labels were collected in Cityscapes~\cite{cityscapes}, erodes and then simplifies the true labels producing controlled masks with various qualities.
In addition, we also compute the estimated number of clicks required to annotate such objects.
This is simulated by counting the number of vertices in the simplified polygon.

\vspace{-3.5mm}
\paragraph{Evaluation Metrics:}
We use two quantitative measures to evaluate our approach in the task of boundary prediction. 
\textbf{1)} We use maximum F-Measure (MF) at optimal dataset scale (ODS),
and \textbf{2)} average precision (AP) for each class.
To evaluate the quality of the improved coarse segmentation masks, we use the intersection-over-union
(IoU) metric.

\subsection{Semantic Boundary Prediction}
\paragraph{Results and Comparisons.}
We first compare the performance of our approach vs current state-of-the-art methods.
Our baselines include CASENet~\cite{yu2017casenet}, and the recently proposed CASENet-S and SEAL~\cite{yu2018seal}.
CASENet-S can be seen as an improved version of CASENet, while SEAL builds on top of CASENet-S and also deals with misaligned labels.

Table~\ref{tb:main} illustrates per category performance in the high quality re-annotated SBD test set. 
Surprisingly, by just introducing the NMS Layer on top of CASENet, our method outperforms SEAL (an approach that deals with misalignment) by more than 1\% in both MF(ODS) and AP.
By combining with active alignment, we can see that the performance is improved even further.
In Table~\ref{cityscapes_sota}, we also evaluate the performance of our method in the  Cityscapes dataset.

While our method outperforms previous state-of-the-art, we emphasize that the main advantage of the proposed approach is its ability of being added on top of any existing architecture such as CASENet, CASENet-S or SEAL.

\begin{figure}[t!]
\vspace{-3mm}
\begin{minipage}{\linewidth}
\centering
\includegraphics[width=0.70\linewidth,height=3cm,trim=30 0 50 32,clip]{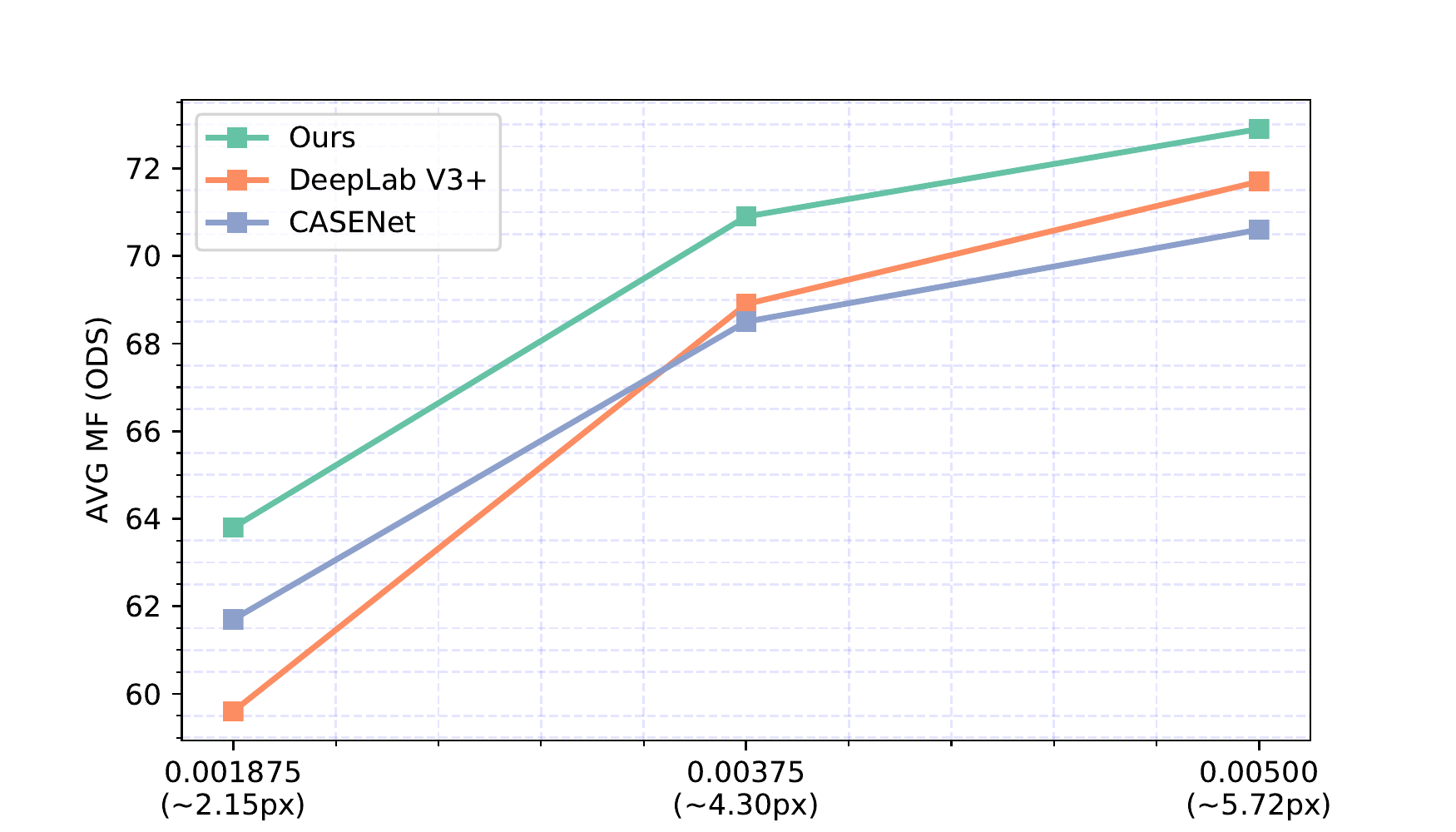} 
\vspace{-3mm}
\caption{Comparison of our boundaries vs those obtained from DeepLab v3+'s segmentation masks. We perform 4.2\% better at the strictest regime.}
\label{fig:vs_state_of_the_art_segm}
\end{minipage}
\end{figure}

\begin{figure}[t!]
\vspace{-3.5mm}
\centering
\begin{minipage}{\linewidth}
\centering
\includegraphics[width=0.72\linewidth,height=3.5cm,trim=15 20 5 5,clip]{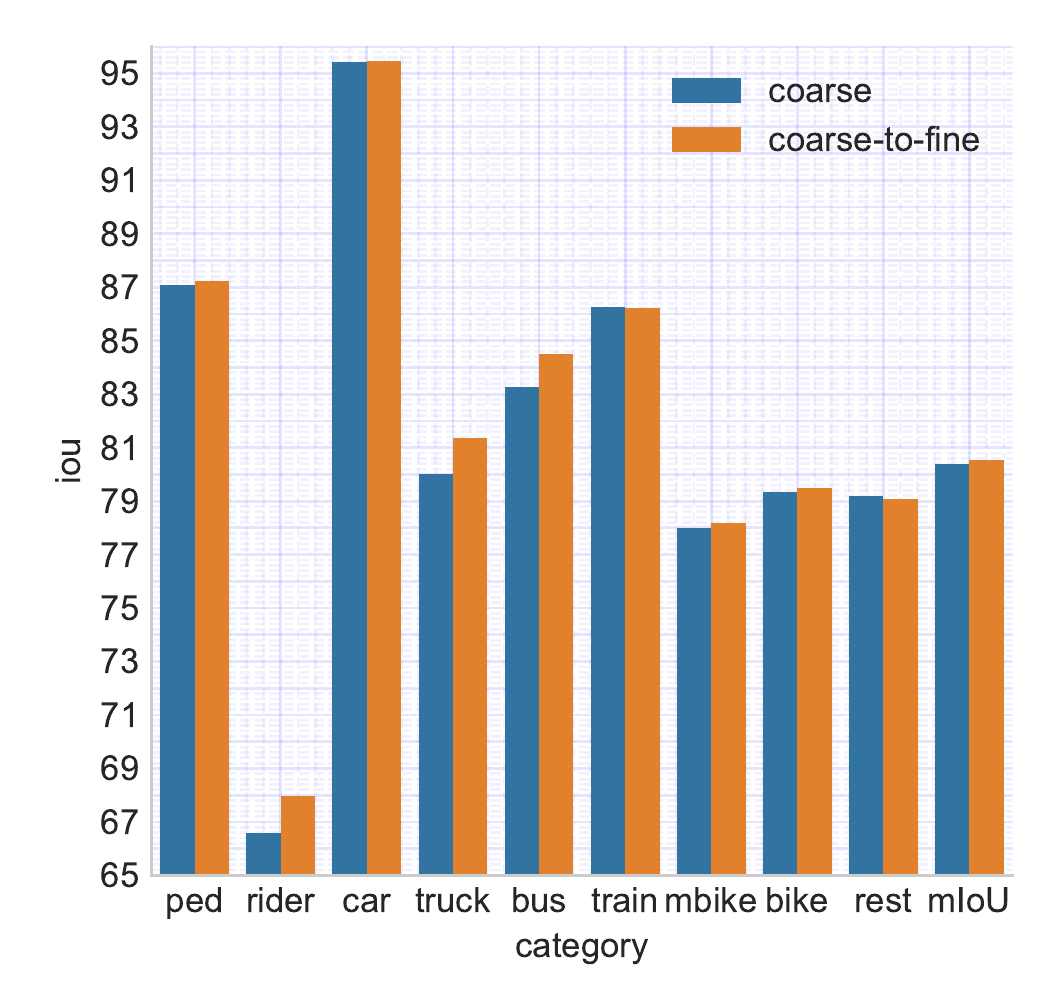} 
\vspace{-3mm}
\caption{Semantic Segmentation on Cityscapes val: Performance of DeepLab V3+ when trained with fine data and 
(blue) vanilla train\_extra set, (orange) our refined data (8 object classes) from train\_extra. We see improvement of more than 1.2 IoU \% in rider, truck and bus.
  }
\label{fig:beating_the_benchmarks}
\end{minipage}
\vspace{-3mm}
\end{figure}

\vspace{-4mm}
\paragraph{Analysis of the Boundary Thinning Layer.}
We evaluate the performance of the NMS and direction loss on the SBD dataset in two different test sets. 
These include the original noisy annotated test set and its re-annotated version from~\cite{yu2018seal}.
The comparison, shown in  Table~\ref{tbl_effect_nms_loss_and_alignment}, highlights the effectiveness of the NMS and direction loss on both test sets. 
In the original test set, our approach improves the performance of CASENet by 3.72\% in terms of MF(ODS) and 17.11\% in terms of AP.
In the high-quality test set, we outperform the baseline by 5.35\% and 18.61\%,  respectively.

\vspace{-4mm}
\paragraph{NMS Loss w/o Edge-NMS:}
We also compare the performance of our method when post-processing is not used at test time.
Table~\ref{tbl_effect_nms_loss_and_alignment} shows that even when the Boundary Thinning Layer is not used during inference, the NMS Loss equally improves the crispness of the raw predictions. 
As such, we can see improvements vs CASENet  of 1.94 \% (MF) and 10.68 \% (AP) in the original dataset, and 1.47 \% (MF) and 10.47 \% (AP) in the re-annotated one.

\vspace{-4mm}
\paragraph{Analysis of Active Alignment.}
We also evaluate the use of our active alignment during training.
To enable a more controlled analysis, we create several noisier versions of the real ground-truth as explained in Sec.~\ref{sec:datasets}. 
Note that given the notion of label error as introduced by~\cite{Zlateski_2018_CVPR}, the original ground-truth is at roughly 4px error, as measured based on the fine (re-annotated) ground-truth. We train our model using active alignment on the noisy training set, and perform evaluation on the high quality test set from~\cite{yu2018seal}.
Results, shown in Table~\ref{tbl_effect_active_alignment}, illustrate the effectiveness of active alignment in both small and extreme noisy conditions.

\begin{figure*}[t!]
\vspace{-2mm}
\centering
\includegraphics[width=\linewidth,trim=0 130 0 10,clip]{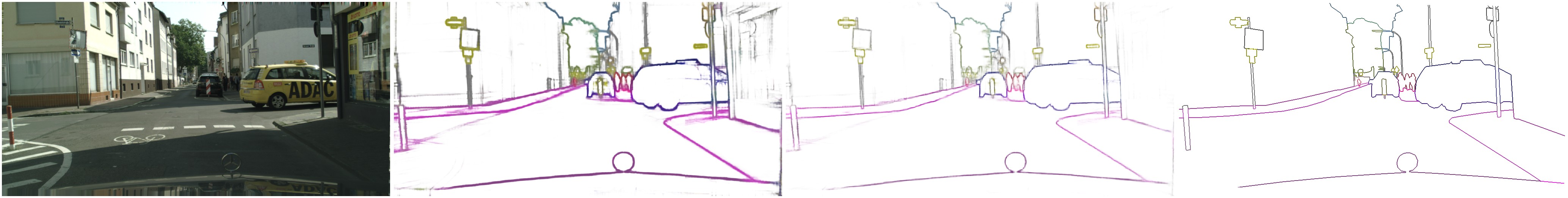}
\includegraphics[width=\linewidth,trim=0 130 0 10,clip]{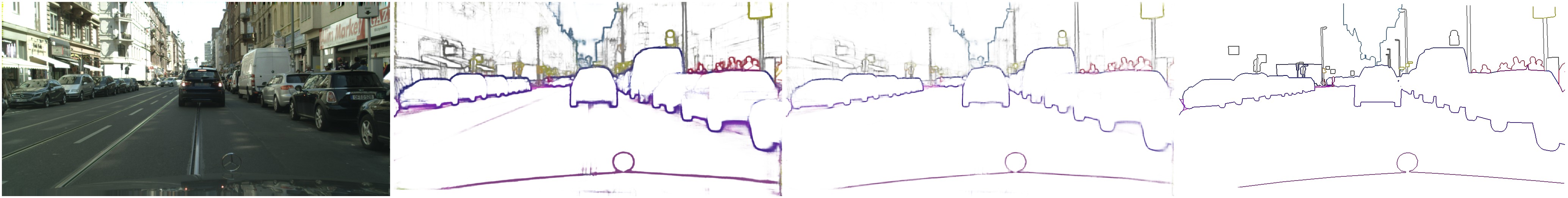}
\includegraphics[width=\linewidth,trim=0 130 0 10,clip]{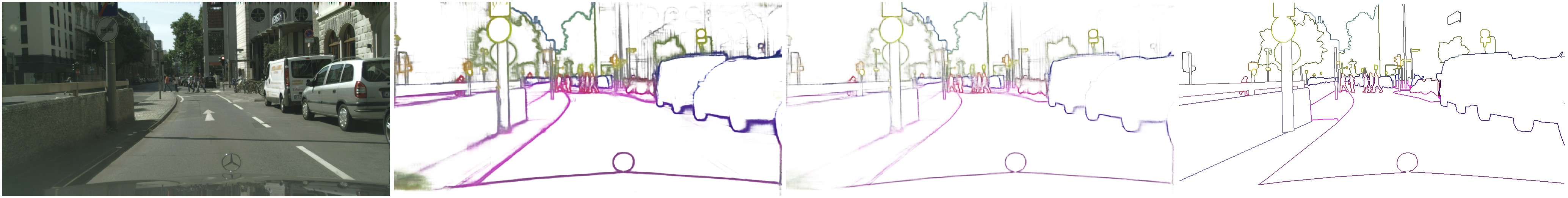}
\includegraphics[width=\linewidth,trim=0 130 0 10,clip]{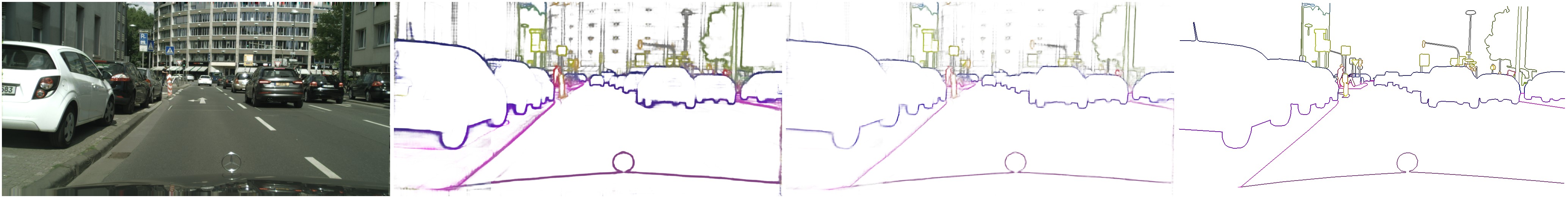}\\[-0mm]
\vspace{-1mm}
\begin{footnotesize}
\centering
\addtolength{\tabcolsep}{38pt}
\begin{tabular}{cccc}
(a) Image & (b) CASENet & (c) Ours & (d) Ground-truth$\ \ $
\end{tabular}
\end{footnotesize}
\vspace{-6.5mm}
\caption{Qualitative Results on the Cityscapes Dataset.}
\label{fig:qualitative_cityscapes}
\vspace{-2mm}
\end{figure*}

\begin{figure*}[t!]
\vspace{-1mm}
\addtolength{\tabcolsep}{-4.0pt}
\begin{tabular}{ccc}
\hspace{-1mm}\includegraphics[height=1.9cm, width=0.33\linewidth,trim=380 225 35 140,clip]{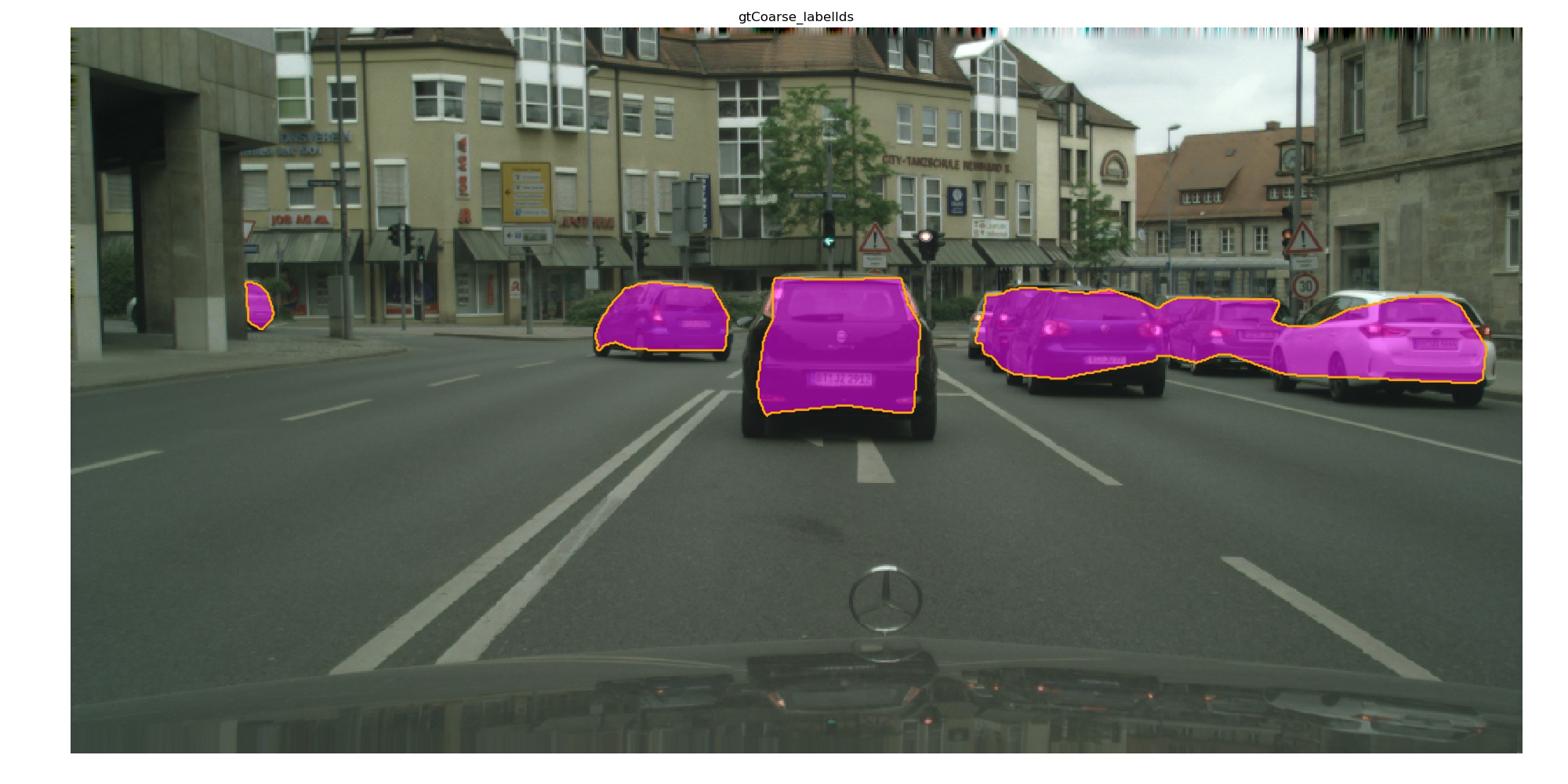} &
\includegraphics[height=1.9cm,width=0.33\linewidth, trim=380 225 35 140,clip]{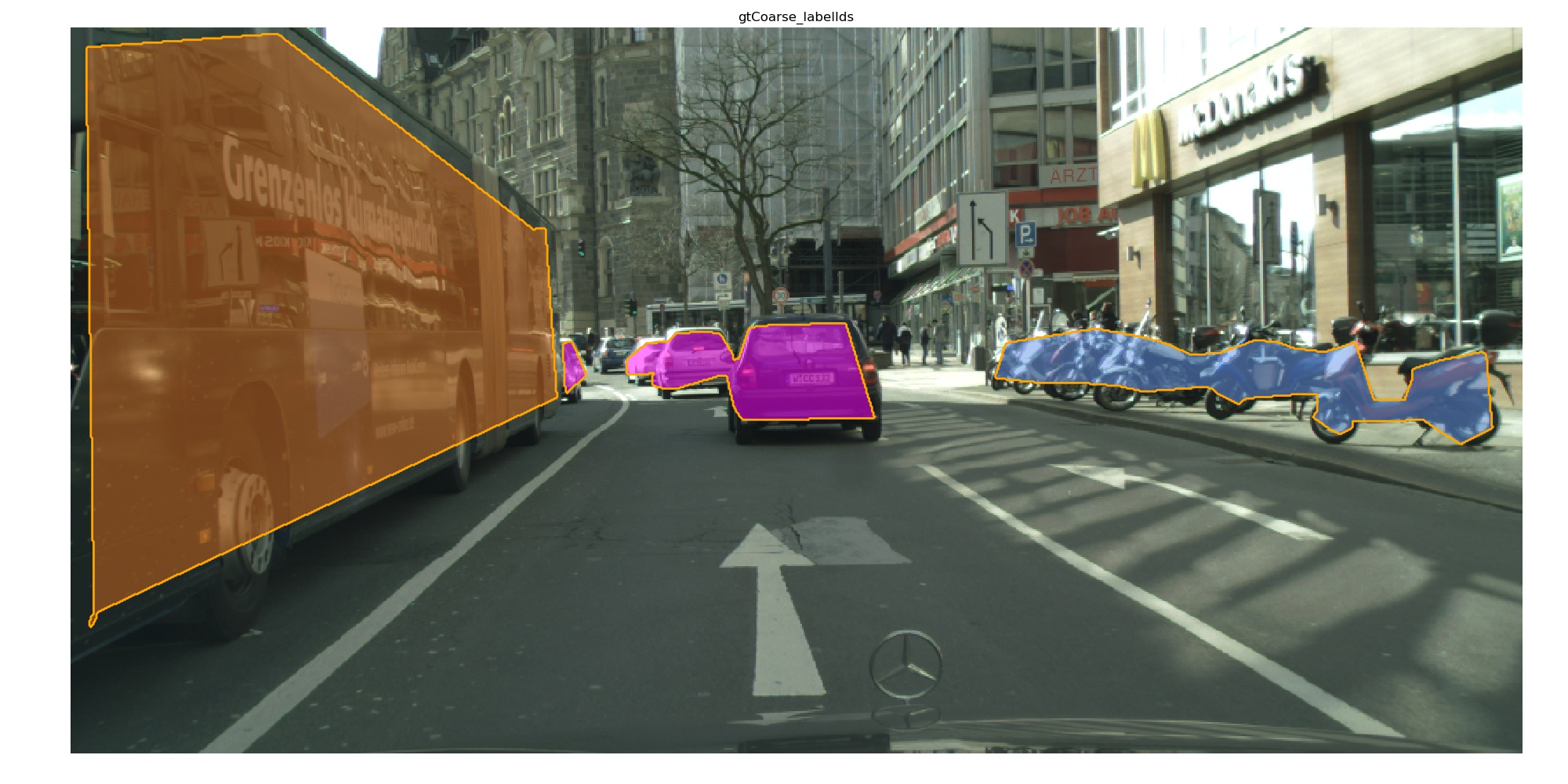}&
\includegraphics[height=1.9cm, width=0.33\linewidth,trim=380 225 35 140,clip]{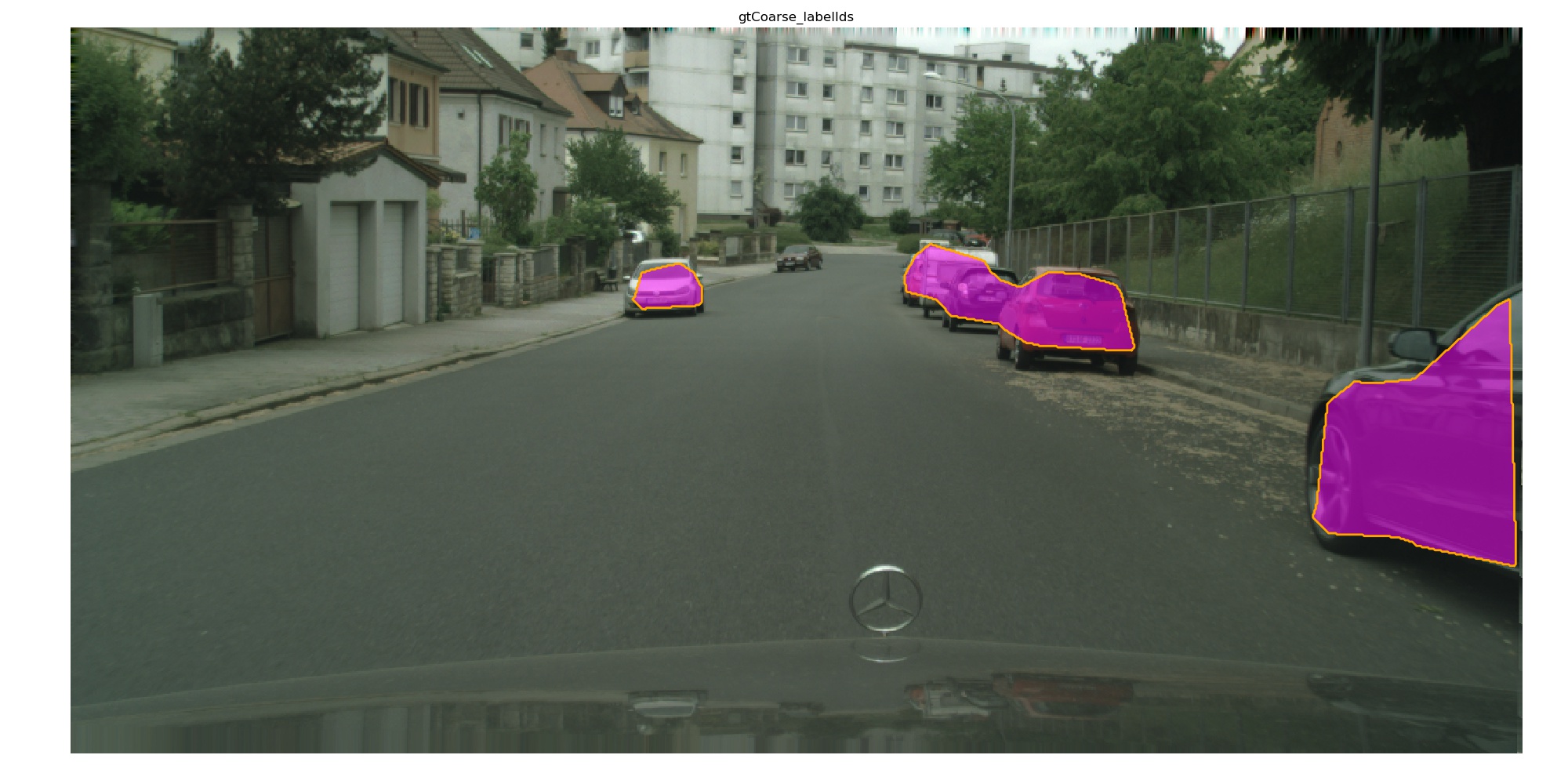}\\[-0.5mm]
\hspace{-1mm}\includegraphics[height=1.9cm, width=0.33\linewidth,trim=380 225 35 140,clip]{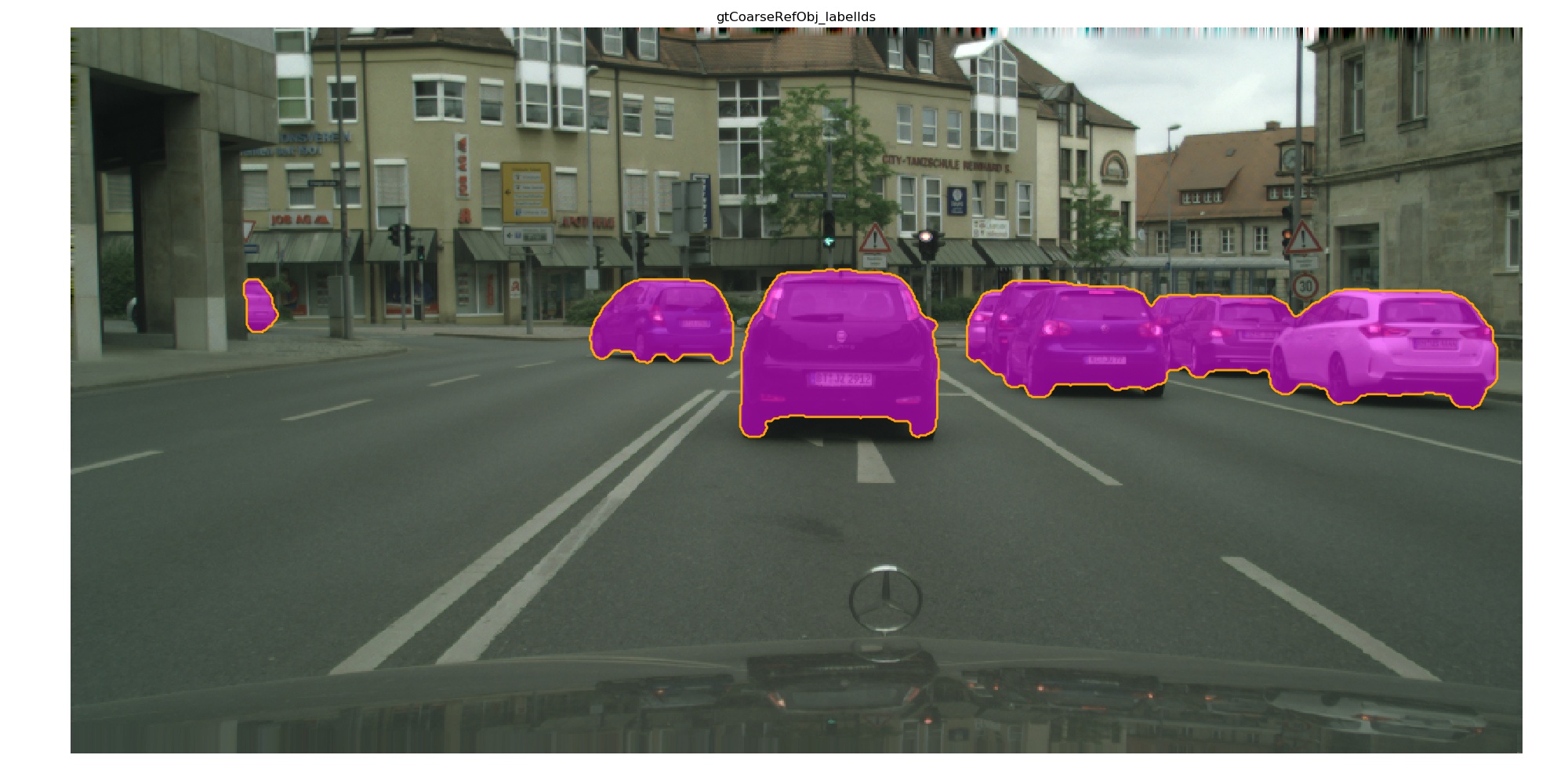}&\includegraphics[height=1.9cm, width=0.33\linewidth,trim=380 225 35 140,clip]{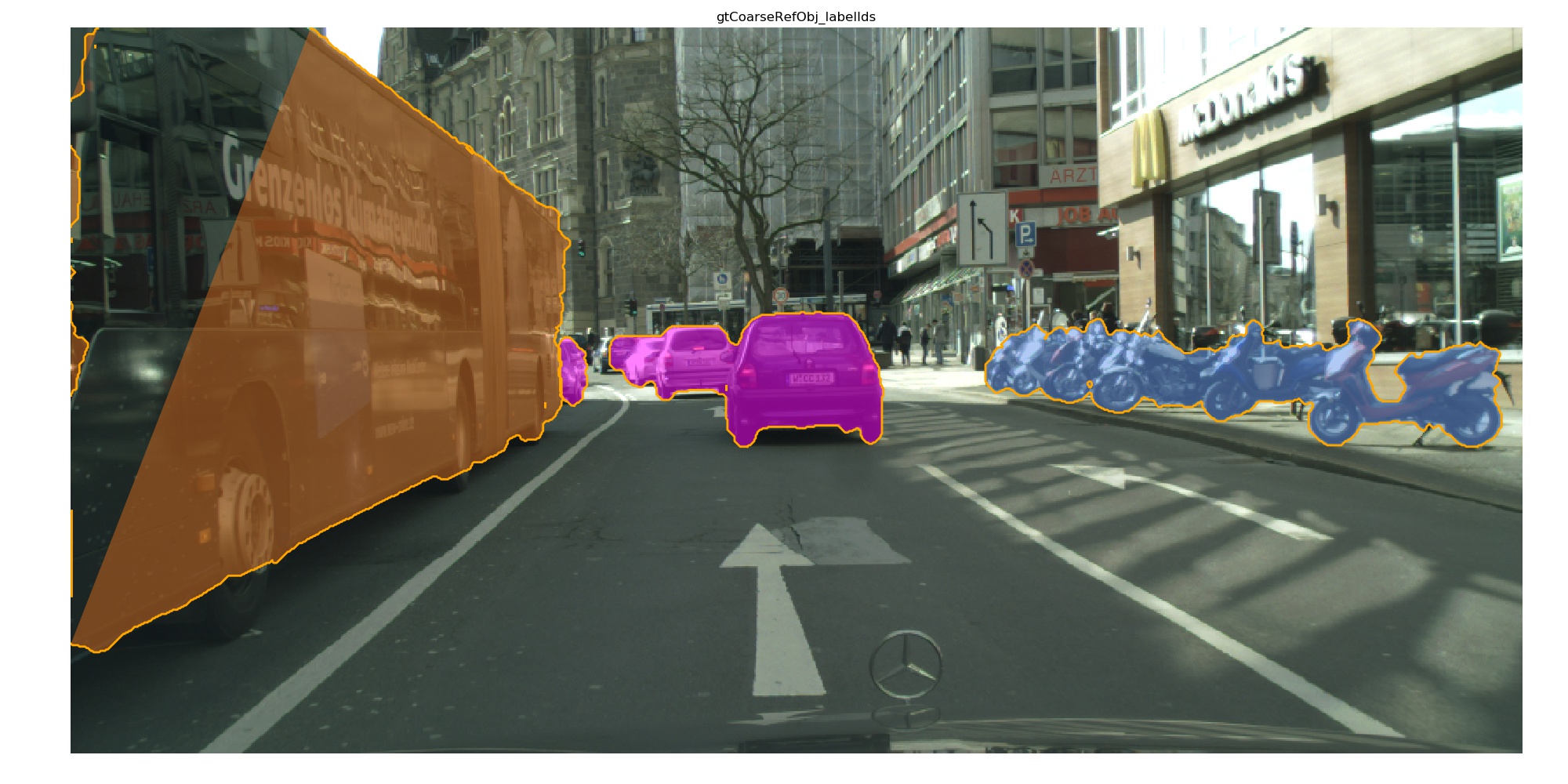}&
\includegraphics[height=1.9cm, width=0.33\linewidth,trim=380 225 35 140,clip]{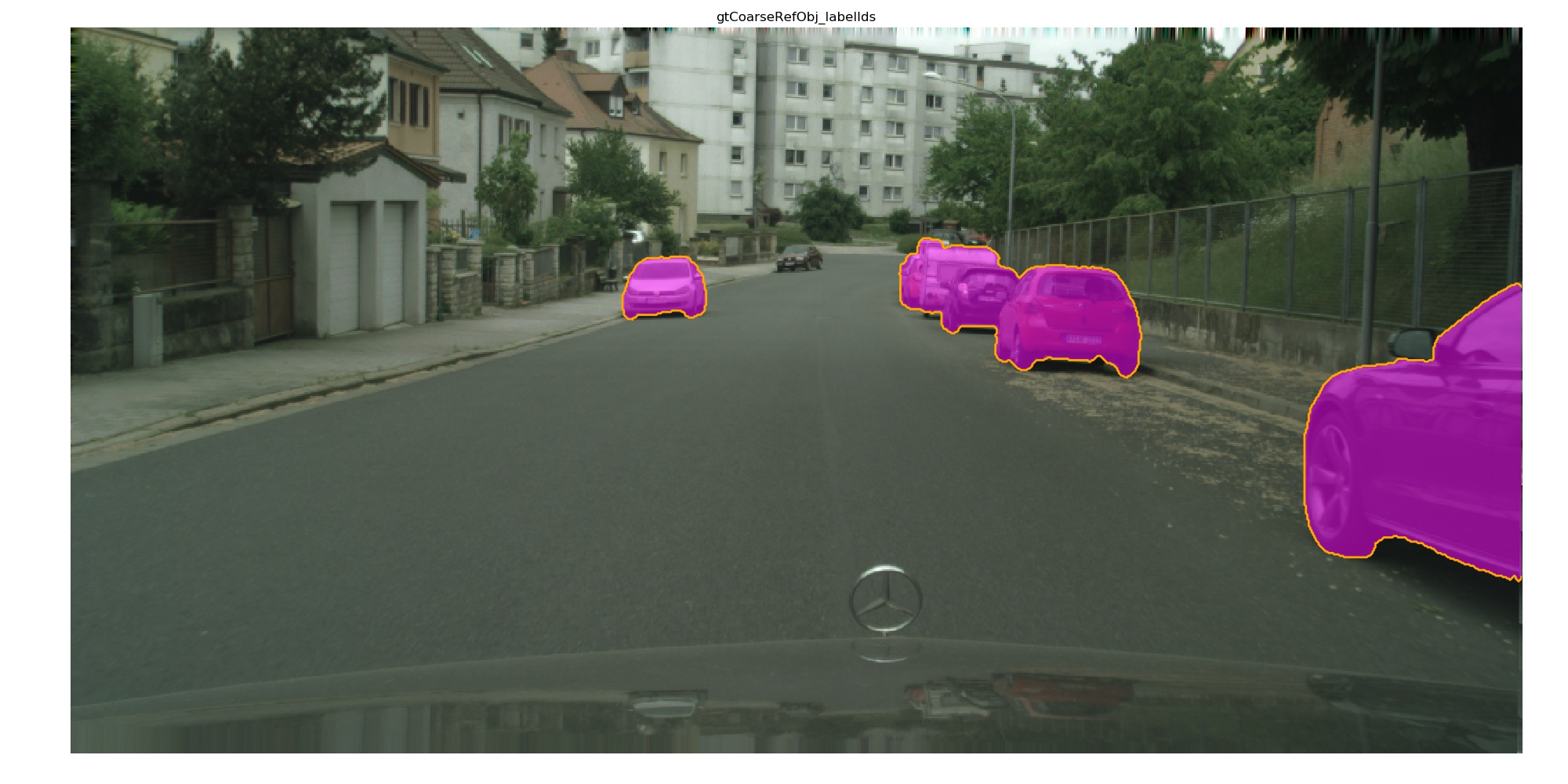}
\end{tabular}
\vspace{-4.5mm}
\caption{Qualitative Results. Coarse-to-Fine on the coarsely annotated Cityscapes train\_extra  set.}
\label{fig:coarse_to_fine}
\vspace{-3.7mm}
\end{figure*}

\vspace{-4mm}
\paragraph{STEAL vs DeepLab-v3~\cite{deeplabv3plus2018}:}
Semantic segmentation can be seen as a dual task to semantic-aware edge detection since the boundaries can easily be extracted from the segmentation masks. 
Therefore, we compare the performance of our approach vs state-of-the-art semantic segmentation networks.
Concretely, we use the implementation of DeepLab V3+ provided by the authors in \cite{deeplabv3plus2018} (78.8 mIoU in the Cityscapes val set), and obtain the edges by computing a sobel filter on the output segmentation masks.
For fairness in evaluation, we set a margin of 5 pixels in the corners and 135 pixels in the bottom of the image.
This removes the ego car and image borders on which DeepLab performs poorly.
The comparison (Fig~\ref{fig:vs_state_of_the_art_segm}), at different matching thresholds, shows that STEAL outperforms DeepLab edges in all evaluation regimes, \eg 4.2\% at $\sim$ 2px thrs. 
This is an impressive result, as DeepLab uses a much more powerful feature extractor than us, \ie Xception 65~\cite{chollet2017xception} vs Resnet101~\cite{he15deepresidual,yu2017casenet}, and further employs a decoder that refines object boundaries~\cite{deeplabv3plus2018}.
The numbers also indicate that the segmentation benchmarks, which compute only region-based metrics (IoU), would benefit by including boundary-related measures. The latter are harder, and better reflect how precise the predictions really are around object boundaries. 

\begin{table}
\vspace{-2mm}
  \centering
  \resizebox{\linewidth}{!}{
\begin{tabular}{|c|c|c|c|c|c}
\hline
Label Quality & 4px error & 8px error  & 16px error & 32px error \\
\hline
Num.Clicks per Image & 70.34 & 44.76  & 26.78 &14.64  \\
Test IoU & 91.22 & 78.95  & 62.20 & 41.31 \\
\hline \hline
GrabCut & 68.74 & 70.32 & 69.76 & 62.82 \\
\hline
Ours(Coarse-to-Fine) IoU & \textbf{92.78} & \textbf{88.16} & \textbf{82.89} & \textbf{76.20} \\
\hline
\end{tabular}}
\centering
\vspace{-3mm}
\caption{{\bf Refining coarse labels on SBD}.  Model is trained on the noisy SBD training set (approx 4px error).
The re-annotated test set is then simplified to simulate coarse data with a given quality (see main text). Score (\%) represents mean over all the 20 object classes.}
\label{tbl_coarse_to_fine_sbd}
\end{table}

\begin{table}
\vspace{-4mm}
  \centering
  \resizebox{\linewidth}{!}{
\begin{tabular}{|c|c|c|c|c|c|c|}
\hline
Label Quality & 4px error & 8px error  & 16px error & 32px error & Real Coarse \\
\hline
Num.Clicks per Image & 175.23  & 95.63    & 49.21  & 27.00  & 98.78\\
Test IoU &74.85   & 53.32   & 33.71 & 19.44 &  48.67    \\
\hline \hline
GrabCut & 26.00  & 28.51    &29.35   & 25.99   &  32.11 \\

\hline
Ours(Coarse-to-Fine) IoU & \textbf{78.93}  &  \textbf{69.21}  & \textbf{58.96} & \textbf{50.35} & \textbf{67.43}     \\
\hline
\end{tabular}}
\centering
\vspace{-3mm}
\caption{{\bf Refining coarse labels on Cityscapes}.  Model trained on fine Cityscapes trainset and used to refine coarse data. Real Coarse corresponds to  coarsely human annotated val set, while x-px error correspond to simulated coarse data. Score (\%) represents mean over all 8 object classes.  }
\label{tbl_coarse_to_fine_cityscapes}
\vspace{-4mm}
\end{table}

\vspace{-4mm}
\paragraph{Qualitative Results}
Fig~\ref{fig:sbd_img2},~\ref{fig:qualitative_cityscapes} show qualitative results of our method on the SBD and Cityscapes datasets, respectively.
We can see how our predictions are crisper than the base network.
In Fig~\ref{fig:sbd_active_align}, we additionally illustrate the true boundaries obtained via active alignment during training. 

\vspace{-2mm}
\subsection{Refining Coarsely Annotated Data}
\vspace{-1mm}

We now evaluate how our learned boundary detection network can be used to refine coarsely annotated data (Sec.~\ref{sec:coarse2fine}). 
We evaluate our approach on both the simulated coarse data (as explained in Sec.~\ref{sec:datasets}), as well as on the ``real'' coarse annotations available in the Cityscapes train\_extra and val sets. For quantitative comparison we use the Cityscapes val set, where we have both fine and coarse annotations. We use the train\_extra set for a qualitative comparison as fine annotations are not available.

 \vspace{-4mm}
\paragraph{Results and Comparisons.} Results of our method are shown in Table~\ref{tbl_coarse_to_fine_sbd} for the SBD dataset. We emphasize that in this experiment the refinement is done using a model trained on noisy data (SBD train set). Table~\ref{tbl_coarse_to_fine_cityscapes}, on the other hand, illustrates the same comparison for the Cityscapes dataset. However, in this case, the model is trained using the finely annotated train set.
In both experiments, we use GrabCut~\cite{Rother2004SIGGRAPH} as a sanity-check baseline. 
For this, we initialize foreground pixels with the coarse mask and run the algorithm at several iterations (1,3,5,10). 
We report the one that gives on average the best score (usually 1).
In our case, we run our method 1 step for the 4px error. For cases, with higher error, we  increase it by 5 starting at 8px error.

\vspace{-4mm}
\paragraph{Qualitative Results.}
We show qualitative results of our approach in Fig~\ref{fig:coarse_to_fine}.
One can observe that by starting from a very coarse segmentation mask, our method is able to obtain very precise refined masks.
We believe that our approach can be introduced in current annotation tools saving considerable amount of annotation time.

\vspace{-4mm}
\paragraph{Better Segmentation.}
We additionally evaluate whether our refined data is truly useful for training.
For this, we refine 8 object classes in the whole train\_extra set  (20K images).
We then train our implementation of DeepLabV3+ with the same set of hyper-parameters with and without refinement in the coarse train\_extra set.
Fig~\ref{fig:beating_the_benchmarks} provides individual performance on the 8 classes vs the rest. 
We see improvement of more than 1.2 IoU\% for rider, truck and bus as well as in the overall mean IoU (80.55 vs 80.37).

\vspace{-2mm}
\section{Conclusion}
\label{sec:conc}
\vspace{-1mm}
 
In this paper, we proposed a simple and effective Thinning Layer and loss that can be used in conjunction with existing boundary detectors. We further introduced a framework that reasons about true object boundaries during training, dealing with the fact that most datasets have noisy annotations. Our experiments show significant improvements over existing approaches on the popular SBD and Cityscapes benchmarks. We evaluated our approach in refining coarsely annotated data with significant noise, showing high tolerance during both training and inference. This lends itself as an efficient way of labeling future datasets, by having annotators only draw coarse, few-click polygons. 

\renewcommand{\baselinestretch}{0.8}
\vspace{-1mm}
\selectfont
\begin{footnotesize}
\paragraph{\footnotesize Acknowledgments.} 
We thank Zhiding Yu for kindly providing the reannotated subset of SBD.  We thank Karan Sapra \& Yi Zhu for sharing their DeepLabV3+ implementation, and Mark Brophy for helpful discussions.

\end{footnotesize}
\renewcommand{\baselinestretch}{1.0}

 \FloatBarrier
{\small
\bibliographystyle{ieee}
\bibliography{egbib}
}

\end{document}